\documentclass{article}

\usepackage[nonatbib, final]{neurips_2025}
\usepackage[numbers]{natbib}
\usepackage{multirow}
\usepackage{subcaption}
\usepackage{wrapfig}
\usepackage{float}

\usepackage[utf8]{inputenc} 
\usepackage[T1]{fontenc}    
\usepackage{hyperref}       
\usepackage{url}            
\usepackage{booktabs}       
\usepackage{amsfonts}       
\usepackage{nicefrac}       
\usepackage{microtype}      
\usepackage[table,xcdraw]{xcolor}

\usepackage{soul}
\usepackage[utf8]{inputenc}
\usepackage{graphicx}
\usepackage{amsmath}
\usepackage{amsthm} 
\usepackage{amsfonts} 
\usepackage{algorithm}
\usepackage{algorithmic}
\usepackage{pdflscape}
\usepackage{multirow}
\usepackage{threeparttable}
\usepackage[table]{xcolor}
\usepackage{tcolorbox}
\usepackage{bbding}
\usepackage{longtable}
\usepackage{natbib}
\usepackage{listings} 
\usepackage[capitalize,noabbrev]{cleveref}

\floatstyle{ruled}
\newfloat{listing}{tb}{lst}{}
\floatname{listing}{Listing}

\newtcolorbox{dialogbox}[1][]{
  arc=4mm,
  colback=gray!3,
  colframe=blue!25!black,
  rounded corners,
  boxrule=0.5pt,
  fonttitle=\sffamily\bfseries,
  coltitle=black,
  colbacktitle=gray!3,
  toptitle=1mm,
  titlerule=0pt,
  bottomtitle=1mm,
  title=#1, 
}
\usepackage{adjustbox}

\definecolor{darkbrown}{HTML}{8B4513}  
\definecolor{navyblue}{HTML}{4682B4}   
\definecolor{darkgreen}{HTML}{228B22}  
\definecolor{purple}{HTML}{9370DB}     
\definecolor{red}{HTML}{CD5C5C}        

\def\ie{\emph{i.e.}}
\def\eg{\emph{e.g.}}
\def\cf{\emph{c.f.}}
\def\etal{{\em et al.}}
\def\etc{{\em etc.}}

\title{Enhancing CVRP Solver through LLM-driven Automatic Heuristic Design}

\author{%
  Zhuoliang Xie\textsuperscript{1} \hspace{1.5em} Fei Liu\textsuperscript{2} \hspace{1.5em} Zhenkun Wang\textsuperscript{1 *}  \hspace{1.5em} Qingfu Zhang\textsuperscript{2} \\
  \textsuperscript{1} Southern University of Science and Technology\\
  \textsuperscript{2} City University of Hong Kong\\
  \texttt{xiezl2025@mail.sustech.edu.cn}, \texttt{fliu36-c@my.cityu.edu.hk},\\
  \texttt{wangzhenkun90@gmail.com}, \texttt{qingfu.zhang@cityu.edu.hk}\\
}

\makeatletter
\renewcommand{\@noticestring}{}
\makeatother

\begin{document}

\maketitle

\begin{abstract}\label{sec:abstract}
  The Capacitated Vehicle Routing Problem (CVRP), a fundamental combinatorial optimization challenge, focuses on optimizing fleet operations under vehicle capacity constraints. 
  While extensively studied in operational research, the NP-hard nature of CVRP continues to pose significant computational challenges, particularly for large-scale instances. 
  This study presents AILS-AHD (Adaptive Iterated Local Search with Automatic Heuristic Design), a novel approach that leverages Large Language Models (LLMs) to revolutionize CVRP solving. 
  Our methodology integrates an evolutionary search framework with LLMs to dynamically generate and optimize ruin heuristics within the AILS method. 
  Additionally, we introduce an LLM-based acceleration mechanism to enhance computational efficiency. 
  Comprehensive experimental evaluations against state-of-the-art solvers, including AILS-II and HGS, demonstrate the superior performance of AILS-AHD across both moderate and large-scale instances. 
  Notably, our approach establishes new best-known solutions for 8 out of 10 instances in the CVRPLib large-scale benchmark, underscoring the potential of LLM-driven heuristic design in advancing the field of vehicle routing optimization. 
\end{abstract}

\section{Introduction}\label{sec:intro}

The Capacitated Vehicle Routing Problem (CVRP) is a critical combinatorial optimization problem that involves managing a fleet of vehicles, each with a specified capacity, to service a set of customers with varying demands. CVRP is of significant practical importance, with applications spanning logistics, production, and transportation~\citep{konstantakopoulos2022vehicleapplication}. The complexity of real-world CVRP scenarios has escalated with the expanding scale of modern businesses and supply systems, presenting substantial challenges for existing CVRP solvers.

Current CVRP solvers fall into three categories: exact methods, deep learning-based neural solvers, and meta-heuristics. Exact methods require extensive computational resources, rendering them impractical for large-scale instances~\citep{baldacci2010exact,laporte1992vehicle}. Neural solvers, while innovative, often encounter challenges related to solution quality and exhibit limitations when confronted with out-of-distribution scenarios. Meta-heuristics, on the other hand, provide satisfactory solutions within a reasonable time and are thus widely adopted in both academic research and industry applications. Notable among these is the Hybrid Genetic Search (HGS)~\citep{vidal2022hybrid}, which leverages a population of solutions and a robust local search strategy. Other significant methods include Slack Induction by String Removals (SISR)~\citep{christiaens2020slack} which uses ruin-and-recreate mechanisms, and Adaptive Iterated Local Search with Path-Relinking (AILS-PR)~\citep{accorsi2021fast} which incorporates techniques like path-relinking and localized optimizations in each iteration. Despite their effectiveness, developing these meta-heuristic approaches often requires extensive expert knowledge and a labor-intensive trial-and-error process. 

Recent advancements have seen Large Language Models (LLMs) being applied in fields such as code generation, mathematical reasoning, and combinatorial optimization~\citep{minaee2024large,romera2024funsearch}. However, the integration of LLMs for enhancing CVRP solvers has yet to reach its full potential. Although recent efforts employ LLMs to enhance the automation of developing routing solvers~\citep{jiang2024unco}, the quality of current solutions remains significantly below the requirements for practical deployment.

In this paper, we present AILS-AHD, an effective adaptive iterated local search framework that combines automatic heuristic design powered by LLMs with an Evolutionary Computation (EC) framework. Through careful analysis of problem characteristics and the AILS framework, we identify and enhance the crucial ruin heuristic in AILS. Our comprehensive experimental evaluation on two standard benchmarks demonstrates the framework's strong performance, achieving 8 new best-known solutions for large-scale CVRPLib instances~\citep{arnold2019efficiently} and one new best-known solution for moderate-scale instances~\citep{uchoa2017new}, representing significant improvements over existing approaches.

Our contributions are summarized as follows:

\begin{itemize}
\item We present AILS-AHD (Adaptive Iterated Local Search with Automated Heuristic Design), an efficient framework that enhances its core ruin method through LLM-driven automatic heuristic design integrated with an evolutionary computation paradigm.
\item We further develop an LLM-based acceleration mechanism incorporating Chain-of-Thought (CoT) technique to efficiently address the computational demands during the evaluation on automated heuristic design.
\item We conduct a comprehensive evaluation against two leading algorithms HGS and AILS-II, showcasing superior solution quality and efficiency across both moderate-scale and large-scale CVRPLib instances. We achieve 8 out of 10 new best-known solutions on large-scale CVRPLib instances and one new best-known solution on moderate-scale instances.
\end{itemize}

\begin{figure*}
    \centering
    \includegraphics[width=0.9\linewidth]{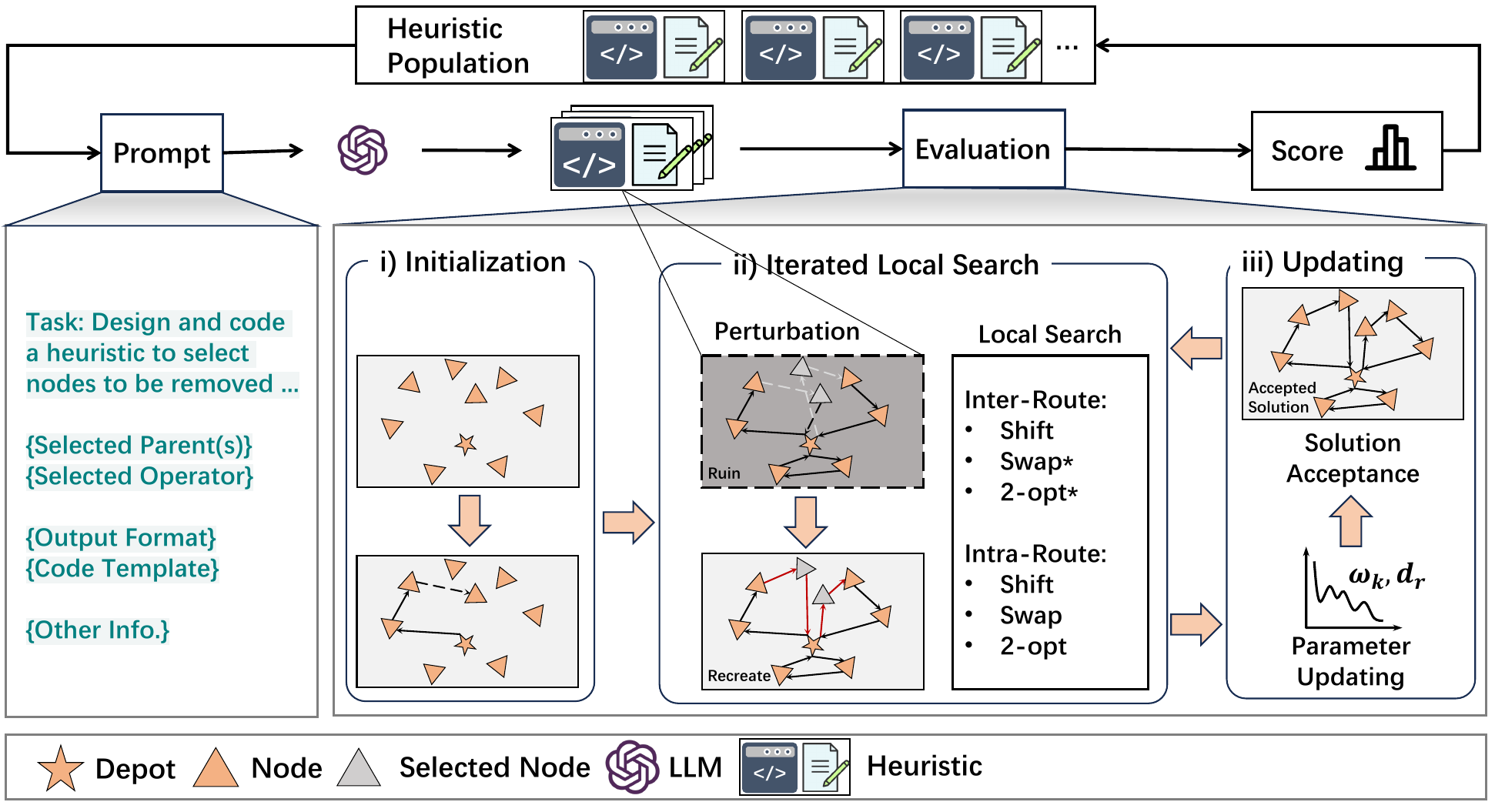}
    \caption{The AILS-AHD pipeline: (1) LLM-driven evolutionary computation generates ruin heuristics; (2) each candidate heuristic is evaluated via AILS, including initialization, iterated local search and updating; (3) the population is updated based on fitness of the heuristic.}
    \label{fig:overall}
\end{figure*}


\section{Problem Description}\label{sec:problem}
The Capacitated Vehicle Routing Problem (CVRP) involves determining optimal routes that originate from a depot, visiting each vertex exactly once, and returning to the depot. Each vertex has a demand that must be served by a vehicle and can only be visited once. The total demand of a route cannot exceed the capacity of its assigned vehicle.

The CVRP can be defined on a complete graph $\mathcal{G} = (\mathcal{V}, \mathcal{E})$, where $\mathcal{V} = \{0, 1, \ldots, n\}$ represents the vertices (nodes), with $n \geq 1$. The vertex $0$ denotes the depot, and $\mathcal{V}_c = \{1, \dots, n\}$ represents the customer set. The demands at these vertices are given by $\mathcal{Q} = \{q_0, q_1, \ldots, q_n\}$, and the edges $\mathcal{E} = \{(i, j) | i, j \in \mathcal{V}, i \neq j\}$ connect all distinct pairs of vertices. The vehicles used are indexed by $\mathcal{M} = \{m_1, \dots, m_l\}$ with $l > 1$, each having the same capacity $c$.

The CVRP is NP-hard. Exact methods, such as branch-and-bound~\citep{boyd2007branch}, quickly become computationally infeasible for large-scale problems. While heuristic algorithms offer a trade-off between computational efficiency and solution quality, they often struggle to maintain this balance as the problem size and complexity increase. The growing scale of real-world applications, characterized by large customer sets and diverse vehicle types, further exacerbates these challenges.

\section{AILS-AHD}\label{sec:ails-ahd}

\subsection{AILS}\label{sec:method}
The backbone Adaptive Iterated Local Search (AILS) consists of three parts: i) Initialization, ii) Iterated local search, and iii) Updating.

\paragraph{Initialization}\label{par:initialize}
In initialization, we generate an initial solution using the greedy insertion method followed by a local search to improve its quality. In the greedy insertion, we first randomly select $n^r$ nodes to construct $n^r$ routes, where $n^r$ is the minimum number of routes required to serve all customers:
\begin{equation}\label{eq:min_routes1}
n^r = \left\lceil \sum_{q_i \in \mathcal{Q}} \frac{q_i}{c} \right\rceil.
\end{equation}
Then, the rest $n-n^r$ nodes are iteratively inserted into the $n^r$ routes. In each iteration, one node is selected randomly and inserted into the position that satisfies the vehicle capacity constraint while minimizing the total cost increment of the current solution. Each solution is then refined through a local search to improve its quality.

\paragraph{Iterated Local Search}\label{par:ils}
We iteratively perform two steps starting from the initial solution to improve the quality. The two steps are (a) perturbation and (b) local search~\citep{ails2022}. 

(a) \textit{Perturbation}: We use ruin-and-recreate~\citep{schrimpf2000rr} for the perturbation step. Ruin involves removing a set number of nodes from the current solution according to specific rules such as random selection and clustering selection. For recreation, the removed nodes are reinserted into the partial solution using the nearest insertion and the best insertion. 


This operator is specifically designed to help the solution escape local optima. To adaptively control the perturbation degree, a parameter $\omega_k$ is determined as the number of nodes to be removed. It is calculated and updated through \equationautorefname~\ref{eq:omega_main}.

After perturbation, the solution may violate constraints, such as exceeding the vehicle capacity limit. To address this and improve the solution quality, an improvement step is performed, including feasibility checking and local search.


(b) \textit{Local Search}: During the improvement phase, diverse local search operators are applied iteratively to achieve both inter-route and intra-route improvements. Inter-route improvement refers to applying these operators between two different routes, while intra-route improvement applies them within a single route. 

For inter-route improvement, the operators include Shift~\citep{osman1993metastrategyshift}, Swap*~\citep{vidal2022hybrid}, and 2-opt*~\citep{potvin1995exchange2opts}. Similarly, for intra-route improvement, the applied operators are Shift, Swap~\citep{osman1993metastrategyswap}, and 2-opt~\citep{lin1965computer2opt}. These operators are executed iteratively to refine the solution.

During this process, a value representing the degree of constraint violations in the current solution is calculated. Solutions with fewer violations (i.e., fewer invalid routes) after applying an operator are accepted. This iterative process continues until a feasible solution is achieved. Once feasibility is ensured, a final local search is performed using the same operators for both inter-route and intra-route improvements. This step ensures thorough optimization and further refines the solution. In the local search phase, the solution with the lowest total cost is accepted.


\paragraph{Updating}\label{par:update}
The updating phase consists of two components: parameter updating and solution acceptance, ensuring the algorithm adapts dynamically during the search process.

(a) \textit{Parameter Updates}: Two key parameters, the perturbation degree $\omega_k$ and the reference distance $d_r$, are updated iteratively to guide the search.

The perturbation degree $\omega_k$, which determines the number of nodes removed during perturbation, is adapted based on the reference distance $d_r$ and the adaptive distance $d_k$:
\begin{equation}
\label{eq:omega_main}
\omega_k = 
\begin{cases}
    \min\left(\omega_k \frac{d_r}{d_k}, n\right), & \text{if } d_r > d_k, \\
    \max\left(\omega_k \frac{d_r}{d_k}, 1\right), & \text{if } d_r \leq d_k,
\end{cases}
\end{equation}
where $n$ is the total number of nodes.

The reference distance $d_r$ is reduced gradually using an exponential decay formula:
\begin{equation}
\label{eq:ref_dis_main}
d_r = d_r \left(\frac{d_{\min}}{d_{\max}}\right)^{\frac{1}{it}},
\end{equation}
where $d_{\min}$ and $d_{\max}$ are the minimum and maximum distances manually set, and $it$ is the current iteration number. This ensures $d_r$ decreases progressively, focusing the search on promising regions.

The adaptive distance $d_k$ is updated as a weighted average of its previous value and the distance between the current solution $s$ and the best solution $s^r$:
\begin{equation}
\label{eq:dk}
d_k = 
\begin{cases}
    \left(1 - \frac{1}{it}\right)d_k + \frac{1}{it}d(s, s^r), & \text{if } it < \gamma, \\
    \left(1 - \frac{1}{\gamma}\right)d_k + \frac{1}{\gamma}d(s, s^r), & \text{if } it \geq \gamma,
\end{cases}
\end{equation}
where $\gamma$ is a parameter that prevents recent updates from having too little influence. The distance $d(s, s^r)$ is computed as $|\mathcal{E} \triangle \mathcal{E}^r|$, which represents the number of different edges between the current solution $s$ and the reference solution $s^r$.


(b) \textit{Solution Acceptance}: To decide whether the solution is accepted as the reference solution, a threshold $\theta$ is calculated:
\begin{equation}
\label{eq:theta_main}
\theta = \underline{f} + \eta (\Bar{f} - \underline{f}),
\end{equation}
where $\underline{f}$ and $\Bar{f}$ are the minimum and average objective values observed so far, and $\eta$ is a dynamic parameter reflecting the quality of solutions. Only solutions with $f(s) < \theta$ are accepted, allowing the algorithm to balance exploration and exploitation effectively. 

\subsection{LLM-driven AHD}\label{sec:ahd}
In the AILS framework, we identify and automatically design the perturbation step in iterated local search. Specifically, we design the heuristic for the ruin operator, i.e., the heuristic determines which nodes are removed given a current solution and features in each iteration.

As illustrated in \figureautorefname~\ref{fig:overall}, we adopt an Evolutionary Computation (EC) framework with LLMs for Automatic Heuristic Design (AHD) of the ruin heuristic. The LLM-driven AHD process maintains and evolves a population of heuristics. In each evolution population, four steps are performed: 1) prompt construction, 2) heuristic generation, 3) evaluation, and 4) population update.

\paragraph{Prompt Construction}
The first step involves constructing a prompt that includes the task description, the required code template format, and other relevant details. The initial parent seed heuristic is manually designed to serve as a starting point. During subsequent iterations, parent heuristics are selected from the population based on a probability distribution defined in \equationautorefname~\ref{eq:select_pro}. Further details about the prompt design, including specific examples and templates, can be found in Appendix.

\begin{equation}\label{eq:select_pro}
    q_i = \frac{f^{s_i}}{\sum\limits_{k \in n^s} f^{s_k}}.
\end{equation}

\paragraph{Heuristic Generation}
Once the prompt is constructed, it is sent to the LLM to generate offspring heuristics. The LLM employs four distinct operators to create new heuristics, each tailored to introduce specific variations or improvements. A detailed description of these operators, including their implementation and usage, is provided in Appendix.


\paragraph{Evaluation}
To evaluate the newly generated heuristics, each is integrated into the original evaluation method by replacing the corresponding component. The heuristic is then tested on a carefully selected set of instances, and its fitness is calculated as the average performance across all instances. This rigorous evaluation process ensures that only high-performing heuristics are retained, maintaining the quality and effectiveness of the population.

\paragraph{Population Update}
After evaluation, the population is updated with the offspring heuristics. To maintain diversity and ensure robust exploration, heuristics with lower fitness are retained in the population.

\subsection{Feature Construction for Heuristic Design}\label{sec:feature}

To efficiently design the ruin heuristic, we construct hybrid features combining problem-specific characteristics and mechanism properties as the heuristic input and output for LLM to design. The key features include:

\begin{itemize}
    \item \textbf{Distance Matrix}: The Euclidean distance matrix $D_{ij}$ between all node pairs $(i,j)$ serves as the spatial foundation, enabling the heuristic to evaluate proximity-based removal priorities. 
    
    \item \textbf{K-Nearest Neighbor List}: For each node $v_i$, we maintain an ordered list $N_K(i)$ of its K closest neighbors. 
    
    \item \textbf{Number of Nodes to Select}: The parameter $\omega_k$ determines the destruction intensity. Larger values promote exploration by removing more nodes, while smaller values favor exploitation through minor perturbations.
    
    \item \textbf{Node Attributes}: Each node $v_i$ is characterized by demand $d_i$ and connectivity with other nodes. For example, nodes with higher demand or critical connectivity roles may be targeted for removal, facilitating more significant improvements during the recreate phase.
    
    \item \textbf{Average Nodes per Route}: The metric $\frac{|\mathcal{V}_c|}{R}$ (where $R$ is the current number of routes) prevents excessive concentration of removals on specific routes.
    
    \item \textbf{Random Seed}: A fixed seed $s$ controls all stochastic operations (e.g., node selection probabilities) during evaluation. This reproducibility mechanism enables fair comparison between different heuristic configurations by eliminating evaluation variance.
\end{itemize}

\textbf{Output Selected Nodes}: The heuristic outputs a set $V_r = \{v_1,...,v_k\}$ of nodes selected for removal, where $k = \omega_k$.

\subsection{Acceleration Mechanism}\label{sec:accelerate}
To address the high computational demand in automatic heuristic design, we propose an LLM-driven acceleration method. Specifically, each generated heuristic is incorporated into a carefully crafted prompt that combines the Chain-of-Thought (CoT) technique~\citep{wei2022chain}, enabling the LLM to assess the heuristic's potential value before actual evaluation. Furthermore, we implement an early stopping mechanism that discards heuristics showing poor performance on initial evaluation instances. The detailed prompt design is provided in Appendix.

\section{Experiment}\label{sec:experiment}

\subsection{AILS-AHD}\label{sec:setting}
\paragraph{Settings}


For the automatic heuristic design phase of the AILS-AHD, we adopt an Evolutionary Computation (EC) framework to maintain the sampled heuristics. The population size is set to 25, and the maximum generation is set to 10 (resulting in 1000 LLM calls). We use 4 EC operators to guide the LLM in heuristic sampling. The detailed operator description is shown in Appendix. For the testing phase, we directly use the top-3 LLM-designed heuristics to integrate into AILS for evaluation.


\begin{figure}
    \centering
    \includegraphics[width=0.7\linewidth]{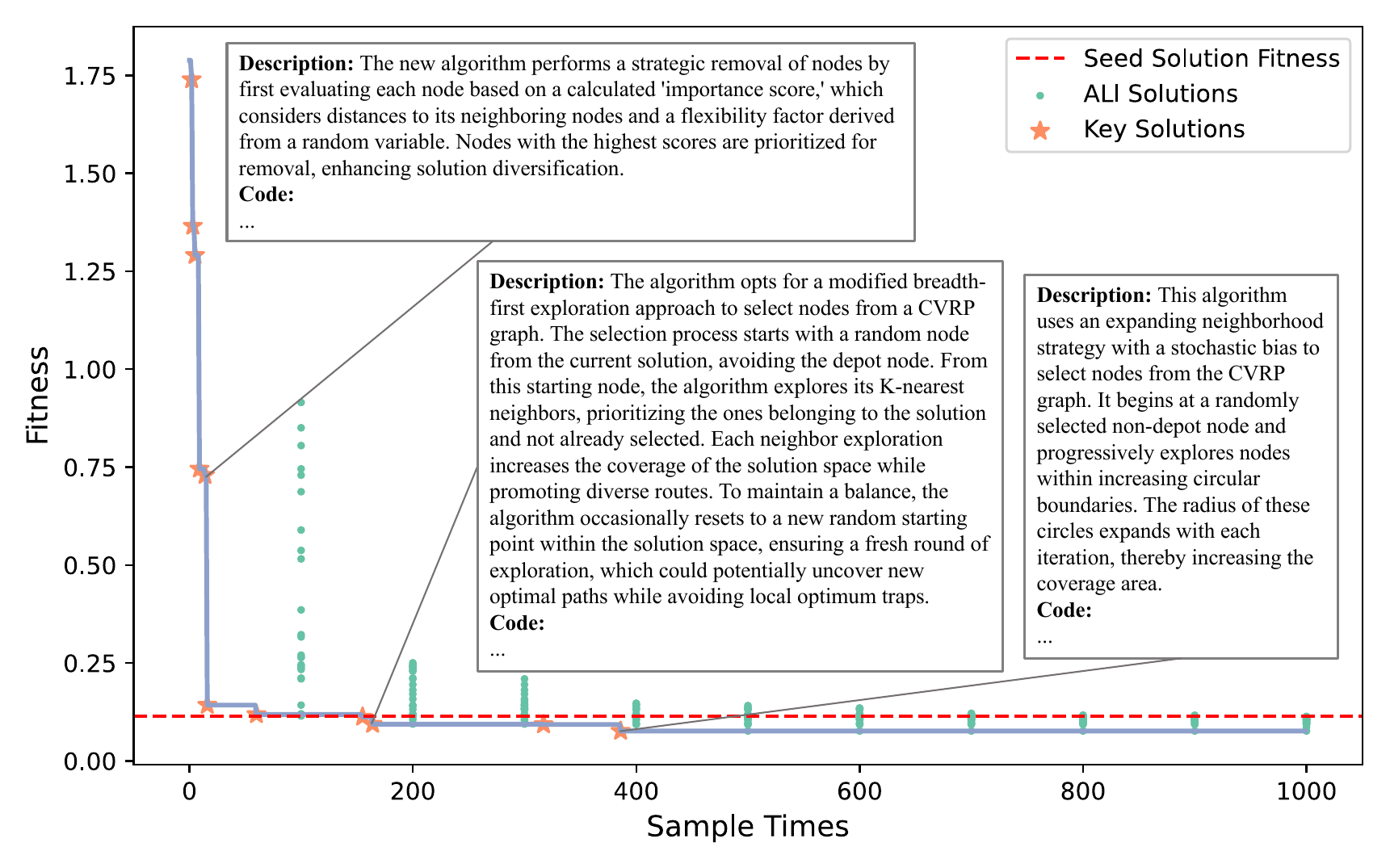}
    \caption{Convergence curve of LLM designed heuristics.}
    \label{fig:ec_llm}
\end{figure}

We conduct experiments using the pre-trained LLM GPT-4o. The specific prompts employed during the experiments are provided in Appendix. To evaluate the heuristics generated through AHD based on LLM process, we utilized 10 problem instances. These are carefully selected taking into consideration the instance scales. For each instance, the evaluation is performed under two different random seeds, resulting in a total of 20 independent instance runs for a single evaluation.

\paragraph{Convergence Process}
As shown in \figureautorefname~\ref{fig:ec_llm}, the best heuristic is identified after approximately 350 samples. To illustrate the convergence process, we list three heuristics along the convergence curve as examples, with the best heuristic highlighted in the orange box.

\paragraph{Designed Heuristics}
Here we list the main idea of the top-3 heuristics. Due to the page limitation, the full code is displayed in Appendix:

\begin{itemize}
    \item \textbf{Expanding Neighborhood Node Selection (EN)}: This is the best-performing heuristic, which operates by first randomly selecting a seed node and then iteratively selecting the remaining nodes from the neighborhood circle of the seed node. It introduces a new dynamic parameter $\lambda$ to dynamically expand the neighborhood radius. The radius of the neighborhood gradually increases in steps of $\lambda$ during the selection process if the number of selected nodes does not meet the total requirement.
    
    \item \textbf{Demand-Driven Node Selection with Distance Decay (DDD)}: It selects nodes based on demand, incorporating a decay factor that reduces distance influence over time. It starts from a random seed node and expands a spiral radius, prioritizing high-demand nodes while ensuring spatial coherence. This approach balances demand efficiency with effective node selection.

    \item \textbf{Probabilistic Node Selection with Frequency Decay (PFD)}: This algorithm selects nodes by initializing a frequency array and iterating until the desired count is reached. It uses a luck threshold to either select nodes sequentially from a random starting point or probabilistically based on proximity and frequency, applying exponential decay to reduce selection likelihood over time.
\end{itemize}

\subsection{Baselines} 
We compare the designed heuristics with two state-of-the-art methods: HGS~\citep{vidal2022hybrid} and AILS-II~\citep{maximo2024ails}. HGS is a representative population-based memetic method that combines a genetic algorithm with local search. AILS-II is the advanced version of AILS, representing a recent single-solution-based method that utilizes an iterated local search framework. 

We also compare two hand-crafted variants of AILS-AHD: AILS-C and AILS-S to further demonstrate the superiority of our automatically designed heuristics. AILS-C uses the K-nearest heuristic, while AILS-S uses the sequence-based heuristic.

\subsection{Benchmark}\label{sec:benchmark}
We use two representative CVRPLib benchmarks:
\begin{itemize}
    \item X Set~\citep{uchoa2017new}: It is the most commonly used CVRP benchmark, which includes 100 moderate-scale instances with node size ranges from 100 to 1,000.
    \item AGS Set~\citep{arnold2019efficiently}: It is a representative large-scale CVRP benchmark. It has 10 large-scale instances extracted from the geometrical data of real-world cities, such as Antwerp and Brussels, with sizes ranging from 3,000 to 30,000. It poses a significant challenge for existing CVRP solvers.
\end{itemize}

All experiments are conducted on 2 Intel(R) Xeon(R) Gold 6254 CPUs. Each instance is executed for $3*n$ seconds for complete convergence according to ~\citet{maximo2024ails}, where $n$ is the number of nodes in the instance. A total of 30 independent runs are performed for each instance.
 

\subsection{Result on Moderate-scale Instances}\label{sec:mod}

We first evaluate different methods on the moderate-scale instances to test their general performance.

\begin{table*}[ht]
\centering
\caption{Results achieved on moderate-scale instances.}
\resizebox{1\textwidth}{!}{
\begin{threeparttable}
\renewcommand{\arraystretch}{1.25}
\begin{tabular}{cccccccccc}
\toprule
Instance & BKS & HGS & AILS-II & AILS-C & AILS-S & Ours-PFD & Ours-DDD & Ours-EN \\
\midrule 
Avg. Gap(\%) (+/-/=) & \textbf{0.00} & 0.109 (40/38/22) & 0.0702 (9/1/90) & 0.0687 (3/2/95) & 0.0804 (30/6/64) & 0.0678 (7/3/90) & 0.0686 (2/3/95) & \cellcolor{gray!30}0.0672 \\ 

\bottomrule
\end{tabular}
\end{threeparttable}
}
\label{tab:res_small_summary}
\end{table*}

As shown in \tableautorefname~\ref{tab:res_small_summary}, BKS represents the Best-Known Solution objectives. Avg. denotes the average objective over 30 runs. The notation ``(+/-/=)'' indicates the statistical significance of the results when comparing Ours-EN with the compared methods. Specifically, ``+'' means Ours-EN performs significantly better than the compared method, ``-'' indicates it performs worse, and ``='' signifies no significant difference between the two methods. The results show that Ours-PFD, Ours-DDD, and Ours-EN consistently outperform HGS, AILS-II, AILS-C, and AILS-S in terms of average performance, where AILS-C and AILS-S are methods based on our AILS framework but replace the ruin heuristic with manually designed ones. Notably, Ours-EN achieves the best average gap of 0.0672\%, demonstrating its superiority over other methods. Detailed results are provided in Appendix.


\begin{figure}[!h]
    \centering
    \includegraphics[width=1\linewidth]{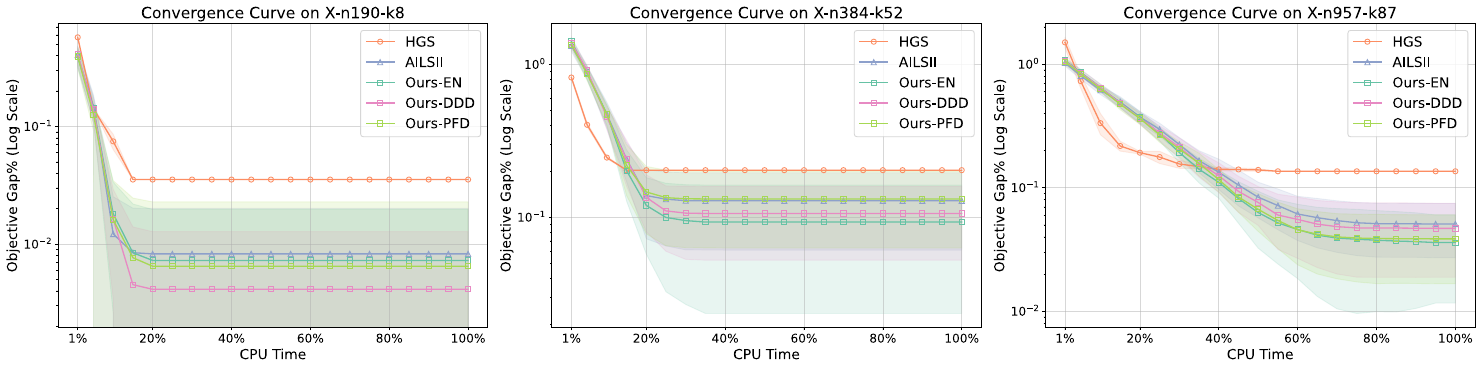}
    \caption{Convergence curve on moderate-scale CVRP instances.}
    \label{fig:con_small}
\end{figure}

As illustrated in \figureautorefname~\ref{fig:con_small}, we also compare the convergence curves across three different instances. The y-axis averaged over 30 independent runs and is shown in log scale. HGS usually demonstrates faster convergence at the beginning with minimal variance. However, it struggles to achieve further improvement beyond that point. In contrast, AILS-II and our designed heuristics show potential for continuous convergence, with the heuristics designed by our method achieving superior results overall. More convergence curves are shown in Appendix.


\subsection{Result on Large-scale Instances.}\label{sec:large}

\begin{figure}[!h]
    \centering
    \begin{subfigure}{0.65\textwidth}
    \resizebox{0.99\textwidth}{!}{
    \begin{threeparttable}
    \begin{tabular}{cccccc}
    \toprule
    Instance & BKS & AILS-II & Ours-EN \\
    \midrule 
    Antwerp1 & \textbf{477261.00}$^a$ & 477536.90 $\pm$ 4.83e+01(+) &  \cellcolor{gray!30}477433.90 $\pm$ 6.49e+01 \\ 
    Antwerp2 & \textbf{291265.00}$^a$ & 291657.00 $\pm$ 1.13e+02(+) &  \cellcolor{gray!30}291430.60 $\pm$ 1.29e+02 \\ 
    Brussels1 & \textbf{501434.00}$^a$ & 501839.10 $\pm$ 3.33e+01(=) &  \cellcolor{gray!30}501786.50 $\pm$ 1.02e+02 \\ 
    Brussels2 & \textbf{344822.00}$^a$ & 345341.00 $\pm$ 1.02e+02(=) &  \cellcolor{gray!30}345321.00 $\pm$ 1.41e+02 \\ 
    Flanders1 & \textbf{7238697.00}$^a$ & \cellcolor{gray!30}7241048.40 $\pm$ 8.90e+02(-) & 7242256.90 $\pm$ 9.31e+02 \\ 
    Flanders2 & \textbf{4365762.00}$^a$ & \cellcolor{gray!30}4370954.10 $\pm$ 1.35e+03(-)  & 4375279.30 $\pm$ 1.87e+03 \\ 
    Ghent1 & \textbf{469482.00}$^a$ & 469752.70 $\pm$ 1.13e+02(=) & \cellcolor{gray!30}469665.20 $\pm$ 1.21e+02 \\ 
    Ghent2 & \textbf{257563.00} & 257826.40 $\pm$ 1.33e+02(=) & \cellcolor{gray!30}257801.40 $\pm$ 1.13e+02 \\ 
    Leuven1 & \textbf{192848.00} & 193025.00 $\pm$ 4.96e+01(+) & \cellcolor{gray!30}192900.10 $\pm$ 5.61e+01 \\ 
    Leuven2 & \textbf{111355.00}$^a$ & 111616.10 $\pm$ 9.74e+01(+) & \cellcolor{gray!30}111497.50 $\pm$ 1.19e+02 \\ 
    \midrule 
    Avg. Gap(\%) (+/-/=) & & 0.1061 (4/2/4) &  \textbf{0.0862} \\ 
    
    \bottomrule
    \end{tabular}
    \begin{tablenotes}
    \footnotesize
    \item[a] Better BKS obtained by Ours-EN.
    \end{tablenotes}
    \end{threeparttable}
}
        \caption{Results achieved on the large-scale instances.}
        \label{tab:res_large}
    \end{subfigure}
    \hfill
     \begin{subfigure}{0.32\textwidth}
        \centering
        \includegraphics[width=\textwidth]{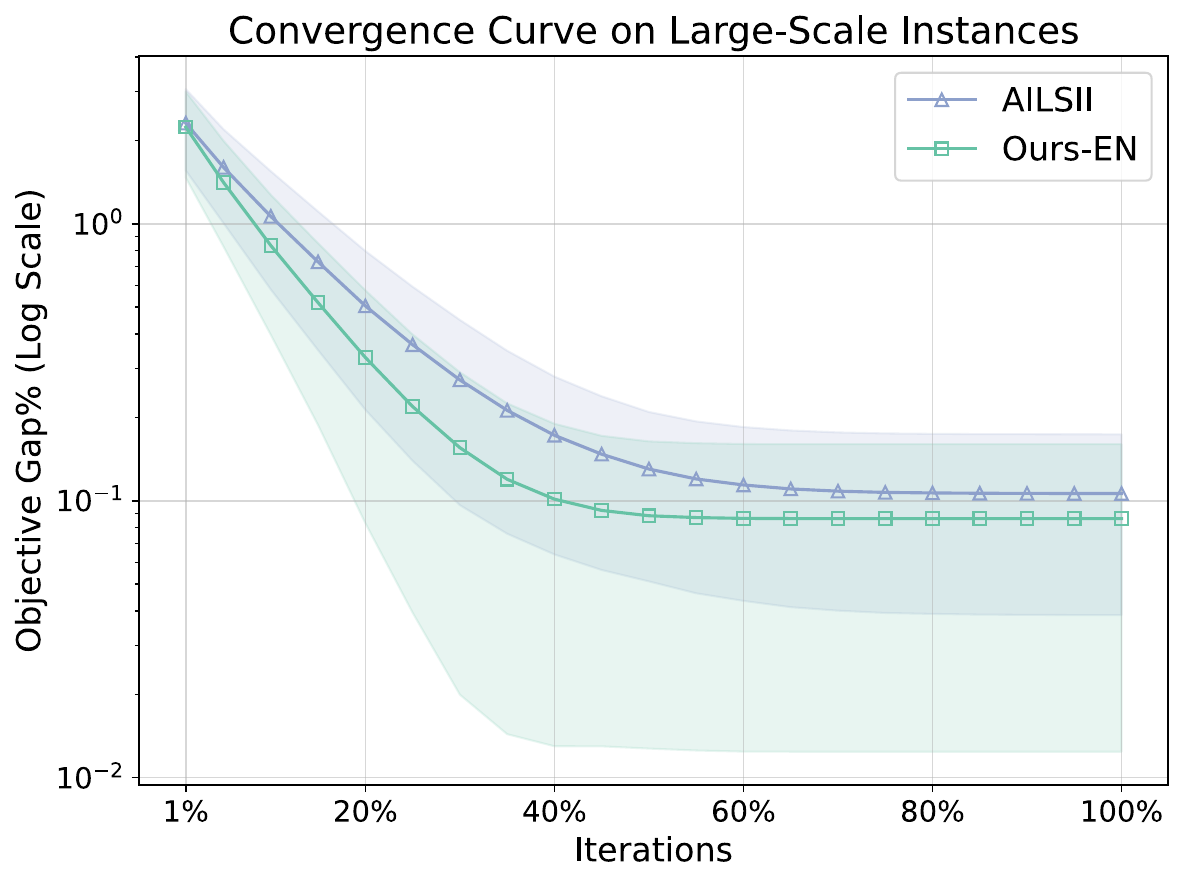}
        \caption{Convergence curve on large-scale CVRP instances.}
    \label{fig:con_large}
    \end{subfigure}
    \caption{Result on large-scale CVRP instances.}
    \label{fig:large_res_all}
\end{figure}

We further evaluate the performance of the best-designed heuristic, Ours-EN, against AILS-II on large-scale instances. The experiments are conducted over 10 independent runs to obtain the average performance, as summarized in \tableautorefname~\ref{tab:res_large}. The results indicate that Ours-EN achieves a lower average gap of 0.0862\%. This demonstrates that Ours-EN consistently outperforms AILS-II in terms of average solution quality across the tested instances.

The average convergence curve is presented in \figureautorefname~\ref{fig:con_large}. The results are averaged over 10 large-scale instances across 10 independent runs. The findings indicate that Ours-EN converges at a faster rate in terms of both CPU time and iterations compared with AILS-II.



We also conduct tests with a runtime of $10 * n$ seconds to search for new best-known solutions for large-scale instances. The best-performing designed heuristic, Ours-EN, is utilized for this evaluation. As a result, we successfully generate \textbf{8 new best-known solutions} out of the 10 instances.

\begin{figure}[!h]
    \centering
    \begin{subfigure}{0.58\textwidth}
        \centering
        \includegraphics[width=\textwidth]{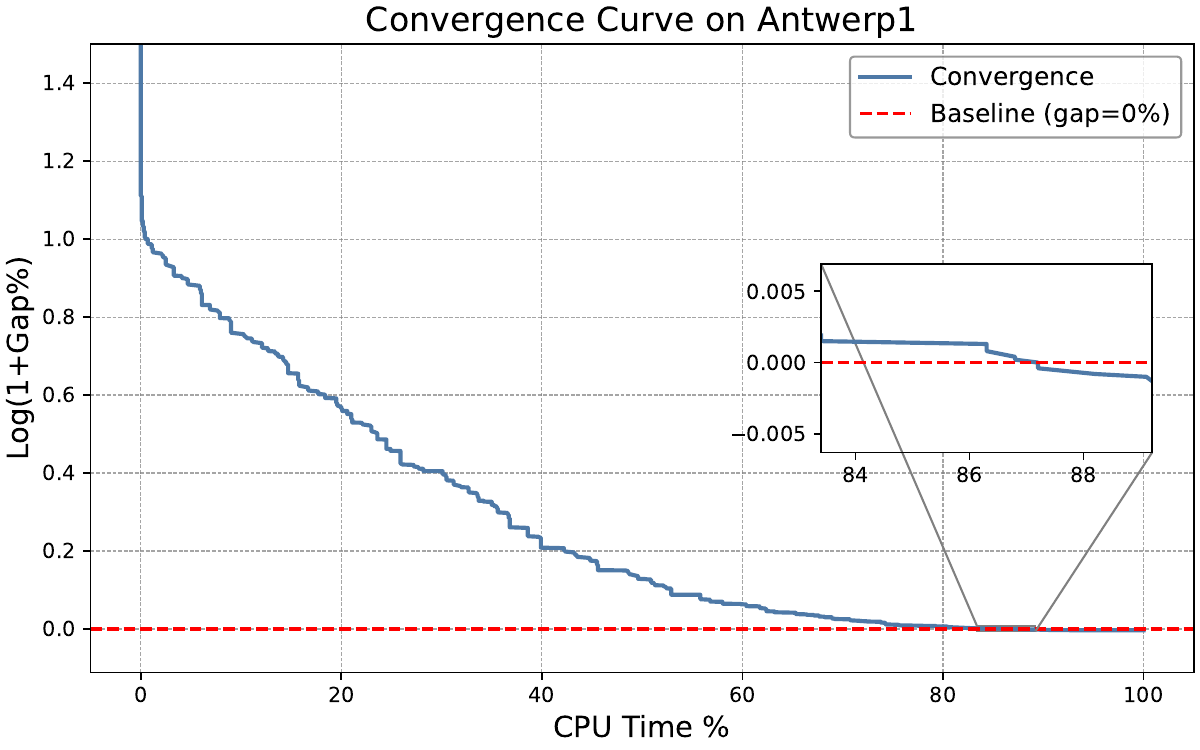}
        \caption{\centering Convergence curve for Antwerp1.}
        \label{fig:antwerp}
    \end{subfigure}
    \hfill
     \begin{subfigure}{0.4\textwidth}
        \centering
        \includegraphics[width=\textwidth]{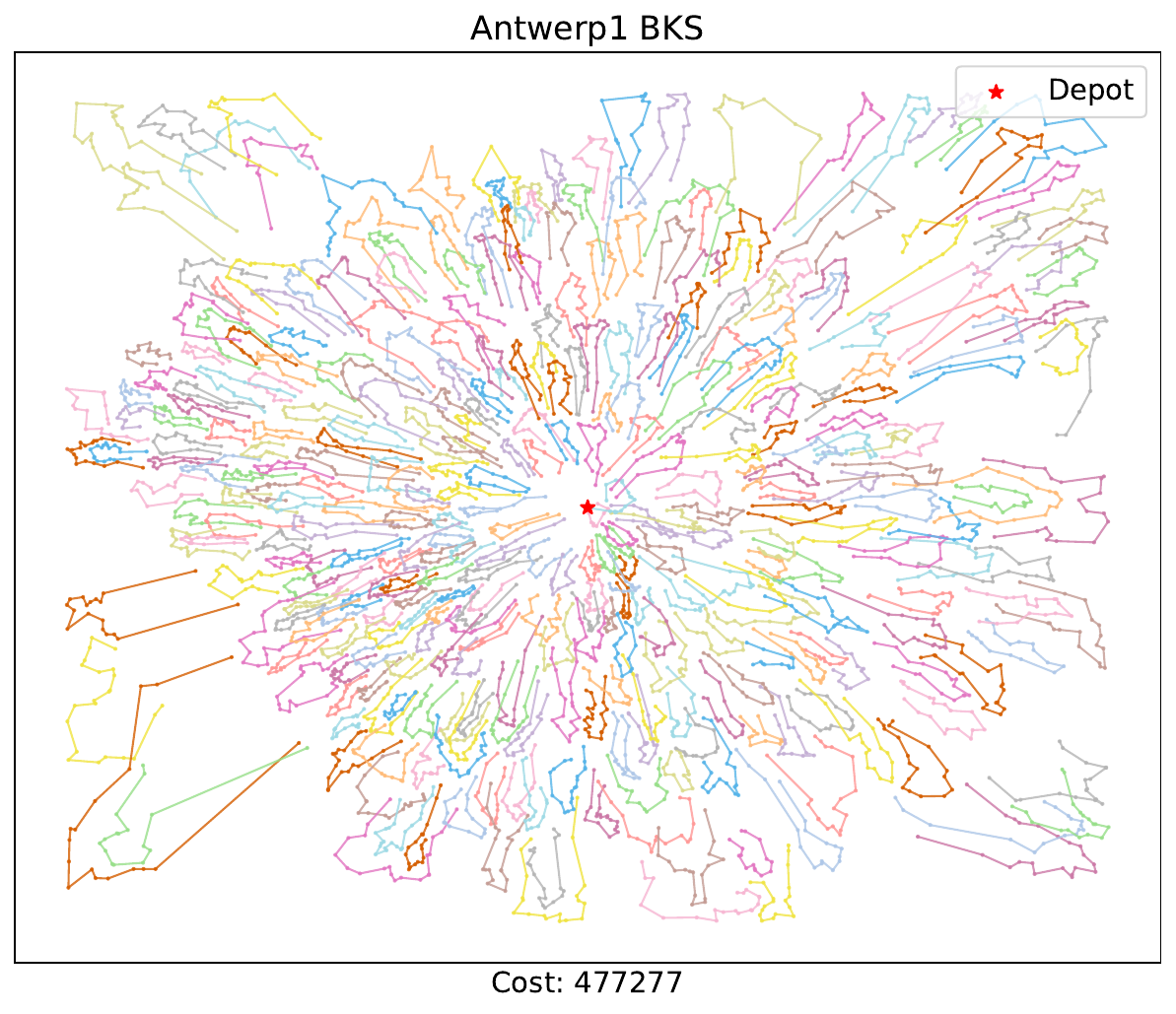}
        \caption{\centering New best-known solution found for Antwerp1.}
        \label{fig:antwerp1}
    \end{subfigure}
    \caption{Convergence curve and new best-known solution found for Antwerp1.}
    \label{fig:large_con_all}
\end{figure}

\figureautorefname~\ref{fig:antwerp} illustrates the convergence curve of Antwerp1. Detailed results and route visualizations of all the new best-known solutions are provided in Appendix.



\subsection{LLM-Driven Acceleration Framework}

Our analysis reveals that in automated heuristic design, most LLM-generated heuristics fail to update the population after evaluation, particularly during later search stages (see \figureautorefname~\ref{fig:ec_llm}). To address this, we conducted experiments on a fixed population comprising 375 heuristics, with results presented in \tableautorefname~\ref{tab:acceleartion}.

\begin{wraptable}[7]{r}{0.5\textwidth}
\vspace{-16pt}
\begin{center}
\renewcommand\arraystretch{1.1}
\caption{Results of LLM-driven acceleration (\%).}
\resizebox{0.5\textwidth}{!}{
\begin{threeparttable}
\renewcommand{\arraystretch}{1.25}
\begin{tabular}{cccccccccc}
\toprule
 & TT & TF & FT & FF & U-R & C-R\\
\midrule 
Vote-1 & 49.3 & 19.7 & 17.9 & 11.2 & 60.5 & 98.1 \\ 
Vote-3 & 51.2 & 17.9 & 19.7 & 9.60 & \textbf{60.8} & 98.4 \\ 
Vote-5 & 51.2 & 18.9 & 20.5 & 8.80 & 60.0 & \textbf{99.4} \\ 
\bottomrule
\end{tabular}
\end{threeparttable}
}
\label{tab:acceleartion}
\end{center}
\end{wraptable}

The first T/F indicates whether the heuristic outperforms the worst population member (warranting evaluation), while the second denotes LLM's judgment. TT and FF represent successful retention of good heuristics and filtering of poor ones, respectively. Our voting mechanism (Vote-n) employs parallel LLM judgments, with U-R measuring acceleration efficiency and C-R showing correct judgment rates. Notably, Vote-3 achieves optimal balance with 60.8\% precision.

We further integrated the Vote-3 acceleration with early stopping mechanisms, demonstrating enhanced efficiency in automated heuristic design (detailed in Appendix).

\subsection{Ablation Study}

\begin{figure}
    \centering
    \includegraphics[width=0.7\linewidth]{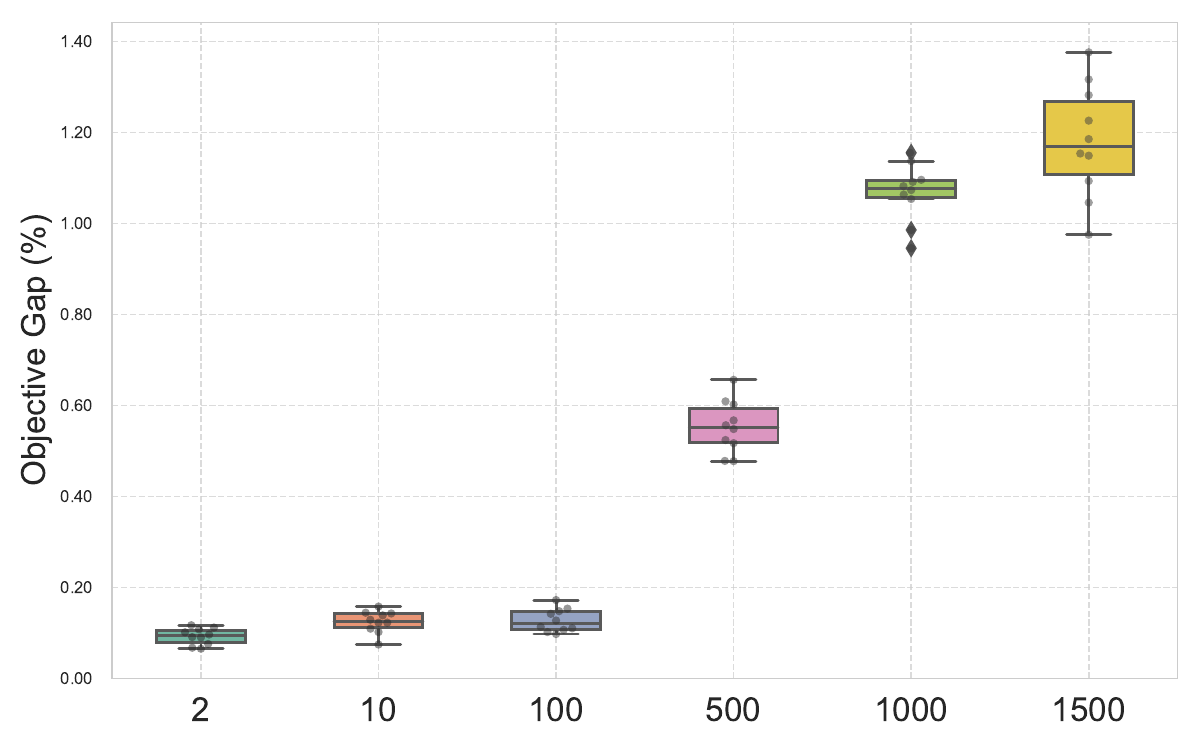}
    \caption{Ablation study on the neighborhood expanding factor.}
    \label{fig:ablation}
\end{figure}

We conduct an ablation study on the neighborhood expanding factor $\lambda$ to analyze its impact on Ours-EN. The factor is set to six different values: 2, 10, 100, 500, 1,000, and 1,500. For each value, we run 10 independent experiments across 10 selected instances. The results are shown in \figureautorefname~\ref{fig:ablation}. The setting with an expanding factor of 2 (the original setting in the LLM-designed heuristic) achieves the smallest objective gap and demonstrates the most stable performance, as indicated by the narrowest box size in the figure. The settings with expanding factors of 10 and 100 exhibit slightly worse performance. However, as the expanding factor increases to 500 and beyond, the performance shows a significant decline, with both worse median values and larger box sizes. The factor of 1,500 yields the worst results, along with the largest box size. These findings suggest that larger neighborhood expansions may lead to instability and poorer results, highlighting the importance of carefully selecting the expanding factor for optimal performance.


\section{Conclusion, Limitation and Future Work}\label{sec:conclusion}

This paper presents AILS-AHD, an efficient framework that integrates Large Language Models (LLMs) with Evolutionary Computation (EC) to automate the design of critical components in meta-heuristics for the Capacitated Vehicle Routing Problem (CVRP). 
To address the inherent randomness in heuristic design for complex problems, we introduce a randomness-alleviation mechanism, enhancing the robustness and reproducibility of the framework. 
By leveraging the generative capabilities of LLMs and the optimization power of EC, 
AILS-AHD significantly reduces the reliance on manual heuristic design while achieving superior performance. 
Experimental results demonstrate the effectiveness of the proposed method, with the top three designed heuristics setting new state-of-the-art results on benchmark datasets. 
Notably, our approach discovers 8 out of 10 new best-known solutions for large-scale instances and one new best-known solution for moderate-scale instances. 
We also propose a simple yet effective LLM-driven acceleration mechanism to reduce the computation demands.
These findings underscore the potential of combining LLMs with EC frameworks to enhance heuristic design in human designed solver. 

\paragraph{Limitation and Future Work}
While AILS-AHD achieves impressive performance on moderate and large scale CVRP instances, it currently relies on a simple AHD framework. In future work, we aim to develop a more efficient AHD framework to further improve the heuristic design efficiency. In addition, we plan to generalize our method to more vehicle routing problem variants.



\clearpage
\appendix

\definecolor{darkbrown}{HTML}{8B4513}  
\definecolor{navyblue}{HTML}{4682B4}   
\definecolor{darkgreen}{HTML}{228B22}  
\definecolor{purple}{HTML}{9370DB}     
\definecolor{red}{HTML}{CD5C5C}        

\def\ie{\emph{i.e.}}
\def\eg{\emph{e.g.}}
\def\cf{\emph{c.f.}}
\def\etal{{\em et al.}}
\def\etc{{\em etc.}}

\title{Enhancing CVRP Solver through LLM-driven Automatic Heuristic Design}


%





\appendix

\section{Related Work}\label{app:related_work}


\subsection{Meta-heuristic for CVRP}

CVRP is an NP-hard problem that requires an extremely large amount of time and computational resources when solved using exact methods, such as Branch and Bound~\citep{boyd2007branch}. While meta-heuristic approaches may not guarantee finding the global optimum, they are capable of producing high-quality approximate solutions within a relatively shorter time and with fewer computational resources. Classic meta-heuristics for solving CVRP can be broadly categorized into two types: single-solution-based methods and population-based methods.

\textbf{Single-solution-based methods} typically maintain only one solution at a time and perform local searches to iteratively improve it. In~\citep{subramanian2013hybrid}, the authors introduced Iterated Local Search with Set Partitioning (ILS-SP), which hybridizes the ILS meta-heuristic with an exact algorithm based on the set partitioning (SP) formulation. The ILS-SP method operates in two stages: local search and perturbation. The local search phase considers both inter-route and intra-route neighborhood structures. The inter-route neighborhood structure employs shift and swap as local search operators, while the intra-route structure utilizes reinsertion, 2-opt, or-opt, and exchange operators. For solution perturbation, ILS applies successive shift and swap movements along with a Randomized Variable Neighborhood Descent (RVND) strategy. During the search process, the best routes generated by ILS are stored, and the exact algorithm subsequently solves the partitioning problem by selecting the optimal set of non-overlapping routes whose union covers the entire set of vertices.

In~\citep{christiaens2020slack}, the authors proposed Slack Induction by String Removals (SISRs), which employs the ruin-and-recreate (R\&R) algorithm with a single ruin method and a single recreate procedure. The ruin method involves removing sequences of adjacent vertices from the same route, while the recreate method reinserts the removed vertices into the lowest-cost positions. Similarly, ~\citep{maximo2021hybrid} introduced Adaptive Iterated Local Search with Path-Relinking (AILS-PR). The AILS component consists of local search and perturbation phases. The local search phase also explores inter-route and intra-route neighborhoods, while the perturbation phase employs a simplified version of the R\&R algorithm with two ruin heuristics (removal by sequence or proximity) and two recreate heuristics (insertion by cost or proximity). Path-Relinking (PR) is used to construct paths between pairs of solutions to identify higher-quality intermediate solutions. However, in~\citep{maximo2024ails}, the authors proposed AILS-II, which removes the PR component due to its high computational cost for large-scale CVRP instances. AILS-II retains most of the AILS-PR structure but introduces slight modifications to the local search operators. Notably, AILS-II incorporates the SWAP* operator proposed in~\citep{vidal2022hybrid} for inter-route local search, which improves upon the standard swap operator by identifying more suitable positions for the selected nodes.

\textbf{Population-based methods} maintain a population of solutions throughout the search process. One of the most well-known population-based methods for solving VRP is Hybrid Genetic Search (HGS)~\citep{uhgs2014a,vidal2022hybrid}. HGS is a memetic algorithm that combines Genetic Algorithms (GA) with local search (referred to as "education" in HGS). The algorithm begins by selecting parent solutions for crossover and mutation based on predefined probabilities to generate offspring. The offspring then undergo a local search process using efficient operators. HGS introduces the SWAP* operator~\citep{vidal2022hybrid}, which identifies better positions for two selected nodes compared to the traditional swap operator, thus enhancing solution quality.

\subsection{LLM for AHD}

Automatic Heuristic Design (AHD) is a systematic approach to creating heuristic functions for solving complex optimization problems. By leveraging computational techniques, AHD aims to automate the design process, reducing the need for manual intervention and enabling the discovery of high-performing heuristics. Recent advancements have explored the integration of Large Language Models (LLMs) into AHD, offering new possibilities for generating and refining heuristics in a scalable and adaptive manner.

Inspired by the process of natural evolution, Evolutionary Computation (EC) serves as a versatile optimization framework~\citep{eiben2015evolutionary}. EC revolves around evolving a population of candidate solutions by simulating biological mechanisms such as genetic variation and natural selection. Over successive generations, these solutions are iteratively refined to achieve optimization. The integration of LLMs into EC has further enhanced its capabilities, enabling the generation of novel heuristics and the exploration of complex solution spaces. For example, some works propose using LLMs to simulate traditional evolutionary operators like mutation and crossover~\citep{meyerson2024language,lehman2023evolution}, while others focus on utilizing LLMs to generate additional contextual or auxiliary information that supports the evolutionary process~\citep{ye2024reevo}. 


While existing approaches leverage the generative power of LLMs and evolutionary optimization to automate heuristic design—providing scalable and adaptive solutions—our focus is on advancing beyond current limitations. We aim not only to enhance state-of-the-art methods but also to outperform human-crafted algorithms, pushing the boundaries of automated algorithm design.


\clearpage

\section{AILS-AHD Settings}\label{app:llm_prompts}

This section presents the detailed prompts used in AILS-AHD in this work.

\subsection{Prompts}

\paragraph{Prompt for Generate New Heuristics}

The prompt for generating new heuristics consists of five key components. \figurename~\ref{fig:prompt_llm} illustrates the overall structure of the prompt provided to the LLM for designing an improved heuristic. The prompt is divided into the following parts:

\begin{itemize}
    \item \textbf{Task Description}: This section provides the LLM with a detailed explanation of the objective and requirements for the strategy to be designed.

    \item \textbf{Heuristic Parent}: This section includes the selected parent heuristic from the current population, chosen through a tournament selection process. It provides information about the historical elite features to guide the iteration process.

    \item \textbf{Operator Selection}: This section specifies an operator to guide the LLM in generating new heuristics. The operator can either encourage diversity or focus on refining existing heuristics.

    \item \textbf{Output Format}: This section instructs the LLM to output the generated heuristic along with its description and the corresponding code in a predefined format. This ensures the output can be easily parsed and utilized.

    \item \textbf{Code Template}: This section provides the LLM with a code template for the designed heuristic. The template ensures that the generated code meets the required syntax and can be evaluated successfully.
\end{itemize}

\paragraph{Seed Heuristic} The process of heuristic initialization starts by adopting the ruin heuristics from the original AILS-II algorithm as the seed heuristics, as illustrated in \figureautorefname~\ref{fig:seed}. These seed heuristics operate by either disrupting the solution through the selection of K-nearest nodes or by choosing successive nodes relative to a designated seed node.

\begin{figure*}[htbp]
    \centering
\begin{minipage}{1\textwidth}
\small
\begin{dialogbox}[EC Operators]
  \textbf{O1}: Please help me create a new heuristic that has a totally different form from the given ones.\\
  \textbf{O2}: Please help me create a new heuristic that has a totally different form from the given ones but can be motivated from them.\\
  \textbf{O3}: Please assist me in creating a new heuristic that has a different form but can be a modified version of the heuristic provided.\\
  \textbf{O4}: Please identify the main heuristic parameters and assist me in creating a new heuristic that has a different parameter settings of the score function provided.\\
\end{dialogbox}
\end{minipage}
\caption{The prompt operators used in AILS-AHD.}
\label{fig:prompt_operators}
\end{figure*}

\begin{figure*}[htbp]

\begin{minipage}{0.99\linewidth}
\begin{lstlisting}[language=java]
|\textbf{<Seed Heuristic Description>}|
"
The algorithm performs a local optimization by randomly selecting a starting node and then iterating through a sequence of its nearest neighboring nodes, optimizing based on their proximity and specific criteria, or alternatively, copying a sequence of nearest nodes directly for further processing.
"

|\textbf{<Code>}|
package Perturbation;

import java.util.Random;
import Solution.newNode;

public class NodeSelector {

    public static int[] nodes_seq(float[][] disMatrix, int[][] nodesKnn, int numberSelect, newNode[] solution, float average_nodes, Random rand){
        int[] selectedNodes = new int[numberSelect];
        int instanceSize = nodesKnn.length;

        int contSizeString;
        int countCandidates = 0;
        double sizeString;
        newNode initialNode;
        newNode node;
        double a = rand.nextDouble();

        if (a > 0.5){
            while(countCandidates<(int)numberSelect)
            {
                sizeString=Math.min(Math.max(1, instanceSize - 1),(int)numberSelect-countCandidates);

                node=solution[rand.nextInt(1, instanceSize)];
                while(!node.nodeBelong)
                    node=solution[rand.nextInt(1, instanceSize)];
                initialNode=node;
                contSizeString=0;
                do
                {
                    contSizeString++;
                    node=node.next;
                    if(node.name==0)
                        node=node.next;

                    selectedNodes[countCandidates++] = node.name;
                    node.nodeBelong=false;
                }
                while(initialNode.name!=node.name&&contSizeString<sizeString);
            }
        }
        else{
            int randomNode = rand.nextInt(1, instanceSize);
            System.arraycopy(nodesKnn[randomNode], 0, selectedNodes, 0, numberSelect);
        }
        return selectedNodes;
    }
}
\end{lstlisting}
\end{minipage}
\caption{Seed heuristic used in AILS-AHD.}
\label{fig:seed}

\end{figure*}

\paragraph{Operators}

Inspired by EoH~\citep{Liu2024EvolutionOH}, we employ four distinct operators to guide the LLM, as depicted in \figureautorefname~\ref{fig:prompt_operators}:

\begin{itemize}
    \item \textbf{O1}: This operator prompts the LLM to generate entirely new strategies by introducing diversity based on the features of the parent strategies. It encourages the exploration of novel solutions that significantly differ from existing ones, promoting a broader search of the solution space.

    \item \textbf{O2}: This operator directs the LLM to refine parent strategies to produce new ones. By leveraging the structure and features of the parent strategies, E2 focuses on creating enhanced versions while preserving some degree of similarity to the original designs.

    \item \textbf{O3}: This operator instructs the LLM to modify a single selected parent heuristic to produce a new heuristic. The modifications are incremental, targeting specific aspects of the parent heuristic to achieve gradual improvements.

    \item \textbf{O4}: This operator concentrates on fine-tuning the parameters of a parent heuristic to generate a new one. It guides the LLM to optimize parameter settings while retaining the core structure of the parent heuristic, aiming to improve overall performance.
\end{itemize}

\begin{figure*}[htbp]
    \centering
\begin{minipage}{1\textwidth}
\small
\begin{dialogbox}[Prompt for Automatic Heuristic Design]
  
  \textcolor {darkbrown}{Task: Design and code a heuristic to select nodes to be removed in a CVRP graph to enhance solution quality. \
The goal is to determine the shortest and most efficient routes from depot node 0, visiting multiple destinations.} \\

  \textcolor{darkgreen}{I have \{number\} heuristic with its code as follows. \\
  No. \{number\} heuristic and the corresponding code are: \\
  \{Heuristic description\} \\
  \{Code\} \\
  ...
  }\\

  \textbf{\{Select one operator\}}\\ 
  
  \textbf{Firstly,} provide the \textcolor{navyblue}{description of your heuristic in braces ``\{\}'' outside the code block.} \\
  \textbf{Next,} implement it in \textcolor{navyblue}{Java as a function implementation} that follows after the description in format: \\
  
  \textcolor{red}{package Perturbation;\\
\\
import Solution.newNode;\\
import java.util.Random;\\
//import other packages if needed;\\
\\
public class NodeSelector \{\\
\\
\hspace*{0.5cm} public static int[] nodes\_seq(float[][] disMatrix, int[][] nodesKnn, int numberSelect, newNode[] nodes, float average\_nodes, Random rand) \{\\
\\
\hspace*{1cm} // disMatrix: Distance matrix of the instance.\\
\hspace*{1cm} // nodesKnn: A sorted int[instanceSize][100] where each element represents nearest nodes id for a given node.\\
\hspace*{1cm} // numberSelect: The number of selected nodes. \\
\hspace*{1cm} // nodes: An array of Node[instanceSize].\\
\hspace*{1.5cm} // Each `Node` object has the following properties:\\
\hspace*{1.5cm} // - `Node.next`: A reference to the next connected Node, or null if there is no connection.\\
\hspace*{1.5cm} // - `Node.prev`: A reference to the previous connected Node, or null if there is no previous connection.\\
\hspace*{1.5cm} // - `Node.nodeBelong`: A boolean, initially true, set to false once the node is selected.\\
\hspace*{1.5cm} // - `Node.name`: An integer representing the node's ID within [0, instanceSize-1]\\
\\
\hspace*{1cm} // average\_nodes: the average number of nodes of each route.
\hspace*{1cm} // node 0 should not be selected.\\
\\
\hspace*{1cm} \texttt{@} Code your heuristic here\\
\\
\hspace*{1cm} return scores;\\
\hspace*{0.5cm} \}\\
\}\\
}

\textcolor{purple}{Do not give additional explanation.}
\end{dialogbox}
\end{minipage}
\caption{Example of prompt engineering used in LLM-driven AHD.}
\label{fig:prompt_llm}
\end{figure*}

\subsection{Evaluation Dataset}

We carefully select 10 diverse instances from the moderate-scale instances of CVRPlib, as proposed by ~\citep{uchoa2017new}. The selected instances are: ``X-n251-k28'', ``X-n256-k16'', ``X-n275-k28'', ``X-n359-k29'', ``X-n411-k19'', ``X-n459-k26'', ``X-n561-k42'', ``X-n613-k62'', ``X-n701-k44'', and ``X-n783-k48''. For each instance, we employ two different random seeds as part of the randomness-alleviation mechanism. This results in a total of 20 runs for evaluating a newly designed heuristic. Each instance is allocated $3 \times n$ seconds to ensure complete convergence, where $n$ represents the number of nodes in the instance. The final evaluation result is calculated as the averaged gap across all instances, providing a robust metric for assessing the performance of the designed heuristic.

\subsection{Early Stopping Mechanism}
During the AHD process, we employ the worst solution in the population as the stopping criterion. If the evaluation result of an instance is worse than this worst solution, the algorithm's evaluation is terminated with a probability defined as:

\begin{equation}
    P = \min(\bar{p}, \max(\underline{p}, \text{gap}))
\end{equation}

where $\underline{p}$ represents the minimum probability threshold, $\bar{p} = \underline{p} + 0.1$ denotes the upper bound, and $\text{gap} \geq \underline{p}$ ensures the probability remains within the specified range.

\clearpage

\section{Designed Heuristics}\label{app:detailed_results}


\paragraph{Ours-PFD}
The heuristic introduces a frequency-based mechanism that balances exploration and exploitation during node selection, as shown in \figureautorefname\ref{fig:pfd}. It utilizes a probabilistic approach to select nodes based on their relative frequency of usage and their proximity to other nodes. A key feature of this method is the dynamic adjustment of the probability threshold, which allows for greater diversity in node selection when randomness is favored. Additionally, an exponential decay mechanism is applied to the frequency component, ensuring that nodes previously selected are gradually deprioritized, promoting a more diverse solution space.

\begin{figure*}[htbp]

\begin{minipage}{1\linewidth}
\begin{lstlisting}[language=java]
|\textbf{<Ours-PFD Description>}|
"
This algorithm adapts node selection by incorporating a node's recent activity and current positional advantage. Nodes are chosen based on the length and quality of sequences they form when linked together, providing an interactive assessment of node potential. By enhancing the score function, the algorithm adopts a decaying influence mechanism, wherein the frequency importance decreases exponentially over time, distributing higher propensity to nodes with immediate route improvement capability. This is designed to provide a balanced effect between maintaining diversity through frequency moderation and focusing on sequence continuity and recent contributions for local optima evasion.
"

|\textbf{<Code>}|
package Perturbation;

import Solution.newNode;
import java.util.Random;

public class NodeSelector {

    public static int[] nodes_seq(float[][] disMatrix, int[][] nodesKnn, int numberSelect, newNode[] nodes, float average_nodes, Random rand) {
        int[] selectedNodes = new int[numberSelect];
        int instanceSize = nodesKnn.length;

        int countCandidates = 0;
        newNode initialNode;
        newNode node;
        float[] frequency = new float[instanceSize];

        for (int i = 0; i < instanceSize; i++) {
            frequency[i] = 0.5f; // Initializes the frequency component with a different starting point
        }

        while (countCandidates < numberSelect) {
            float luck = rand.nextFloat();
            if (luck > 0.8) { // Adjusted luck threshold
                node = nodes[rand.nextInt(1, instanceSize)];
                while (!node.nodeBelong)
                    node = nodes[rand.nextInt(1, instanceSize)];

                initialNode = node;
                int sequenceLength = Math.min(instanceSize, (int)(numberSelect - countCandidates));
                int contSizeString = 0;
\end{lstlisting}
\end{minipage}
\caption{Heuristic of Ours-PFD.}
\label{fig:pfd}

\end{figure*}

\begin{figure*}[htbp]
\begin{minipage}{1\linewidth}
\begin{lstlisting}[language=java]
                do {
                    contSizeString++;
                    node = node.next;
                    if (node.name == 0) {
                        node = node.next;
                    }
                    if (countCandidates < numberSelect) {
                        selectedNodes[countCandidates++] = node.name;
                        node.nodeBelong = false;
                    }
                } while (initialNode.name != node.name && contSizeString < sequenceLength);
            } else {
                int randomNode = rand.nextInt(1, instanceSize);
                for (int i = 0; i < numberSelect && countCandidates < numberSelect; i++) {
                    int candidateNode = nodesKnn[randomNode][i];
                    if (nodes[candidateNode].nodeBelong) {
                        if (rand.nextFloat() < 1 / (frequency[candidateNode] + Math.exp(-i))) { // Modified score function
                            selectedNodes[countCandidates++] = candidateNode;
                            nodes[candidateNode].nodeBelong = false;
                            frequency[candidateNode] *= 0.95; // Exponential frequency decay
                        }
                    }
                }
            }
        }
        return selectedNodes;
    }
}
\end{lstlisting}
\end{minipage}
\caption{Heuristic of Ours-PFD -- continue.}
\label{fig:pfd-1}

\end{figure*}

\paragraph{Ours-DDD}
The heuristic employs a demand-driven strategy, integrating a priority-based mechanism to select nodes, as shown in \figureautorefname\ref{fig:ddd}. This approach utilizes a spiral expansion model, where nodes within an expanding radius around a randomly chosen starting node are evaluated for selection. A weighted priority score is calculated for each node, combining factors such as demand, distance, and a decay factor to prioritize nodes that contribute more significantly to the solution. By leveraging a priority queue, this method ensures that the most promising nodes are selected iteratively, while also maintaining diversity in the solution by introducing randomness into the selection process.

\begin{figure*}[htbp]

\begin{minipage}{0.99\linewidth}
\begin{lstlisting}[language=java]
|\textbf{<Ours-DDD Description>}|
"
This modified algorithm emphasizes demand but introduces a decay factor to the distance influence over time, providing flexibility in node selection. Initially, the distance has a reduced effect, but as the selection progresses, nodes far from the initial node are less favored. This encourages earlier inclusion of nodes that heavily contribute to balanced demand while maintaining a spatially coherent path, promoting partitions that are demand-dominant yet spatially efficient.
"

|\textbf{<Code>}|
package Perturbation;

import Solution.newNode;
import java.util.Random;
import java.util.PriorityQueue;
import java.util.Comparator;

public class NodeSelector {

    public static int[] nodes_seq(float[][] disMatrix, int[][] nodesKnn, int numberSelect, newNode[] nodes, float average_nodes, Random rand) {

        int[] selectedNodes = new int[numberSelect];
        int instanceSize = nodes.length;
        boolean[] selected = new boolean[instanceSize];

        // Prevent selecting the depot
        selected[0] = true;

        int count = 0;
        float spiralRadius = 0.0f;
        float incrementRadius = 1.0f;

        // Priority Queue to manage node selection based on weighted score
        PriorityQueue<WeightedNode> priorityQueue = new PriorityQueue<>(Comparator.comparingDouble(wn -> -wn.priority));

        // Start from a random initial node
        newNode startNode = nodes[rand.nextInt(1, instanceSize)];
        while (!startNode.nodeBelong) {
            startNode = nodes[rand.nextInt(1, instanceSize)];
        }
\end{lstlisting}
\end{minipage}
\caption{Heuristic of Ours-DDD.}
\label{fig:ddd}

\end{figure*}

\begin{figure*}[htbp]

\begin{minipage}{0.99\linewidth}
\begin{lstlisting}[language=java]

        while (count < numberSelect) {
            spiralRadius += incrementRadius;

            // Add nodes to the priority queue for the current spiral epoch
            for (newNode node : nodes) {
                if (node.nodeBelong && node.name != 0 && !selected[node.name]) {
                    float distance = disMatrix[startNode.name][node.name];
                    if (distance <= spiralRadius) {
                        float demand = node.demand;
                        float decayFactor = (float)Math.pow(0.9, spiralRadius); // Introduce a decay factor
                        float weightedPriority = (demand * decayFactor) / (distance + 1) + rand.nextFloat();
                        priorityQueue.add(new WeightedNode(node.name, weightedPriority));
                    }
                }
            }

            // Select nodes from the priority queue
            while (!priorityQueue.isEmpty() && count < numberSelect) {
                WeightedNode wn = priorityQueue.poll();
                if (!selected[wn.nodeIndex]) {
                    selectedNodes[count++] = wn.nodeIndex;
                    selected[wn.nodeIndex] = true;
                    nodes[wn.nodeIndex].nodeBelong = false;
                }
            }
        }

        return selectedNodes;
    }

    static class WeightedNode {
        int nodeIndex;
        double priority;

        WeightedNode(int index, double priority) {
            this.nodeIndex = index;
            this.priority = priority;
        }
    }
}
\end{lstlisting}
\end{minipage}
\caption{Heuristic of Ours-DDD -- continue.}
\label{fig:ddd-1}

\end{figure*}

\paragraph{Ours-EN}
The heuristic adopts an expanding neighborhood node selection process, where nodes are chosen based on their proximity to a starting node within an expanding circular radius, as shown in \figureautorefname\ref{fig:en}. This method ensures that nodes closer to the starting point are prioritized, while also considering the need to explore a broader area by gradually increasing the radius. The heuristic avoids revisiting previously selected nodes and prevents the inclusion of depot nodes, ensuring a valid and efficient solution. By focusing on spatial proximity and systematic exploration, \textbf{Ours-EN} provides a balanced approach to node selection that promotes convergence while maintaining diversity.

\begin{figure*}[htbp]

\begin{minipage}{0.99\linewidth}
\begin{lstlisting}[language=java]
|\textbf{<Ours-EN Description>}|
"
This algorithm employs an expanding concentric circle strategy with a stochastic bias to select nodes from the CVRP graph. Starting from a randomly chosen non-depot node, it progressively investigates nodes within growing circular boundaries. The radius of these circles increases with every iteration, enhancing the coverage area. This approach provides a expanding strategy search to avoid local optima while effectively covering the solution space.
"

|\textbf{<Code>}|
package Perturbation;

import Solution.newNode;
import java.util.Random;

public class NodeSelector {

    public static int[] nodes_seq(float[][] disMatrix, int[][] nodesKnn, int numberSelect, newNode[] nodes, float average_nodes, Random rand) {

        int[] selectedNodes = new int[numberSelect];
        int instanceSize = nodes.length;
        boolean[] selected = new boolean[instanceSize];

        // Prevent selecting the depot
        selected[0] = true;

        int count = 0;
        float circleRadius = 0.0f;
        float increaseFactor = 2.0f;

        // Starting from a random non-depot node
        newNode startNode = nodes[rand.nextInt(1, instanceSize)];
        while (!startNode.nodeBelong) {
            startNode = nodes[rand.nextInt(1, instanceSize)];
        }

        while (count < numberSelect) {
            circleRadius += increaseFactor;

            for (newNode node : nodes) {
                if (node.nodeBelong && node.name != 0 && !selected[node.name]) {
                    float distance = disMatrix[startNode.name][node.name];
                    if (distance <= circleRadius && count < numberSelect) {
                        selectedNodes[count++] = node.name;
                        selected[node.name] = true;
                        nodes[node.name].nodeBelong = false;
                    }
                }
            }
        }

        return selectedNodes;
    }
}
\end{lstlisting}
\end{minipage}
\caption{Heuristic of Ours-EN.}
\label{fig:en}

\end{figure*}

\clearpage

\section{More Experimental Results}\label{app:more_results}

\subsection{More Convergence Curves}\label{app:more_con}

Due to the fact that solutions for small-scale instances are mostly proven to be optimal, in this subsection, we first present the convergence curves of the largest 12 instances within the moderate-scale instances. Overall, our designed heuristics demonstrate superior convergence performance compared to the benchmark methods.

\subsection{Detailed Results on Moderate-scale Instances}

In this subsection, we present detailed results for every instance within the moderate-scale dataset. Surprisingly, one of our methods, Ours-PDF, has successfully discovered a new best-known solution for the instance X-n1001-k43.

\begin{figure*}
    \centering
    \includegraphics[width=0.9\linewidth]{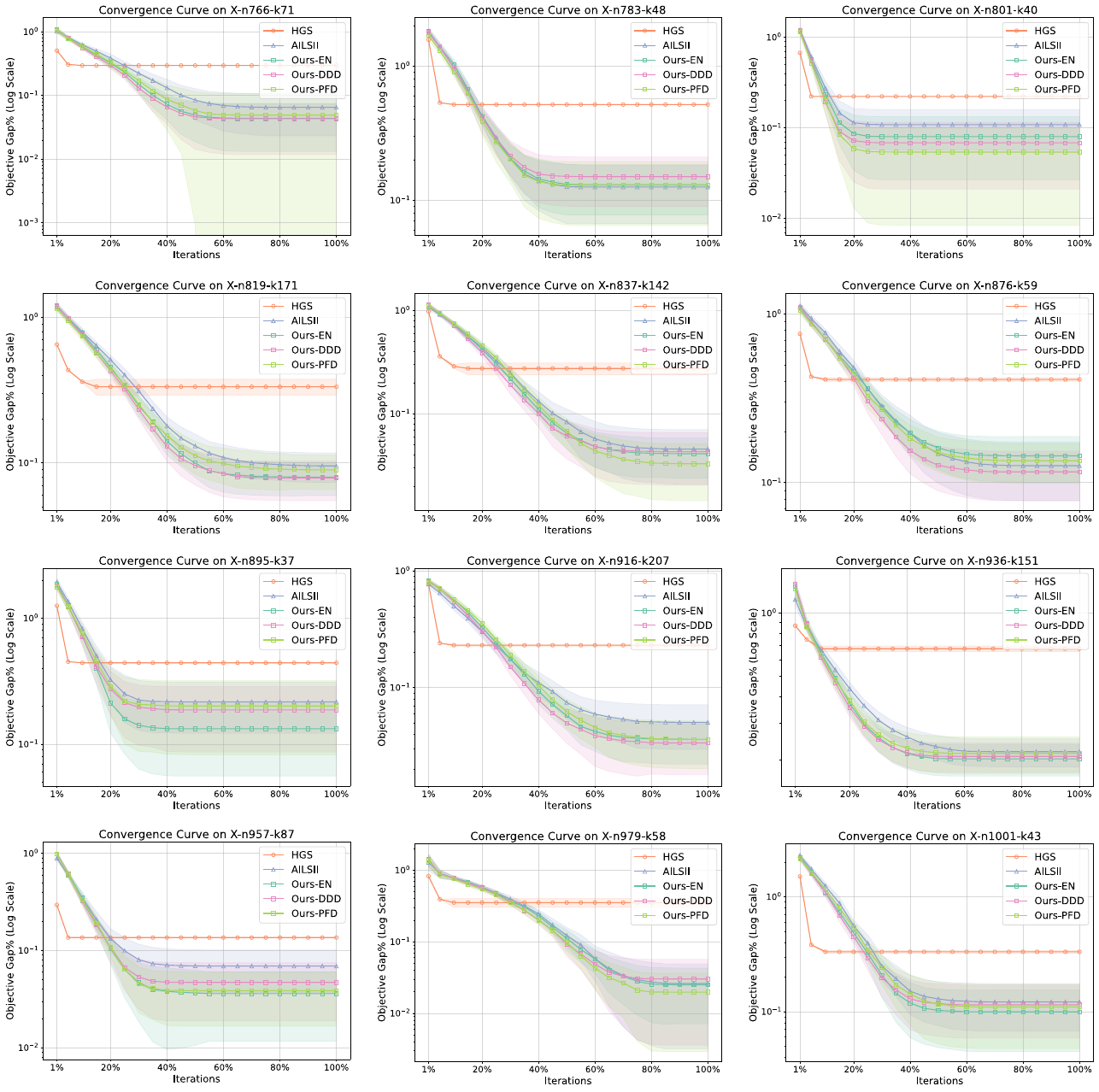}
    \caption{More convergence curves on moderate-scale instances.}
    \label{fig:enter-label}
\end{figure*}

\begin{table*}[ht]
\centering
\resizebox{0.9\textwidth}{!}{
\begin{threeparttable}
\begin{tabular}{cccccccc}
\toprule
Instance & BKS & \multicolumn{2}{c}{HGS} & \multicolumn{2}{c}{AILS-II} & \multicolumn{2}{c}{Ours-EN} \\
\cmidrule(lr){3-4} \cmidrule(lr){5-6} \cmidrule(lr){7-8} \\
 &  & Best & Avg.(\%) (+/-/=) & Best & Avg.(\%) (+/-/=) & Best & Avg.(\%) \\
\midrule 
X-n101-k25 & \textbf{27591.0000} & 27591.0000 & 27591.0000 $\pm$ 0.00e+00(=) & 27591.0000 & 27591.0000 $\pm$ 0.00e+00(=) & 27591.0000 & 27591.0000 $\pm$ 0.00e+00 \\ 
X-n106-k14 & \textbf{26362.0000} & 26376.0000 & 26376.0000 $\pm$ 0.00e+00(+) & 26362.0000 & 26362.0000 $\pm$ 0.00e+00(=) & 26362.0000 & 26362.0000 $\pm$ 0.00e+00 \\ 
X-n110-k13 & \textbf{14971.0000} & 14971.0000 & 14971.0000 $\pm$ 0.00e+00(=) & 14971.0000 & 14971.0000 $\pm$ 0.00e+00(=) & 14971.0000 & 14971.0000 $\pm$ 0.00e+00 \\ 
X-n115-k10 & \textbf{12747.0000} & 12747.0000 & 12747.0000 $\pm$ 0.00e+00(=) & 12747.0000 & 12747.0000 $\pm$ 0.00e+00(=) & 12747.0000 & 12747.0000 $\pm$ 0.00e+00 \\ 
X-n120-k6 & \textbf{13332.0000} & 13332.0000 & 13332.0000 $\pm$ 0.00e+00(=) & 13332.0000 & 13332.0000 $\pm$ 0.00e+00(=) & 13332.0000 & 13332.0000 $\pm$ 0.00e+00 \\ 
X-n125-k30 & \textbf{55539.0000} & 55539.0000 & 55539.0000 $\pm$ 0.00e+00(=) & 55539.0000 & 55539.0000 $\pm$ 0.00e+00(=) & 55539.0000 & 55539.0000 $\pm$ 0.00e+00 \\ 
X-n129-k18 & \textbf{28940.0000} & 28940.0000 & 28940.0000 $\pm$ 0.00e+00(-) & 28940.0000 & 28941.8667 $\pm$ 4.06e+00(=) & 28940.0000 & 28941.7333 $\pm$ 3.21e+00 \\ 
X-n134-k13 & \textbf{10916.0000} & 10916.0000 & 10916.0000 $\pm$ 0.00e+00(=) & 10916.0000 & 10916.0000 $\pm$ 0.00e+00(=) & 10916.0000 & 10916.0000 $\pm$ 0.00e+00 \\ 
X-n139-k10 & \textbf{13590.0000} & 13590.0000 & 13590.0000 $\pm$ 0.00e+00(=) & 13590.0000 & 13590.0000 $\pm$ 0.00e+00(=) & 13590.0000 & 13590.0000 $\pm$ 0.00e+00 \\ 
X-n143-k7 & \textbf{15700.0000} & 15700.0000 & 15700.0000 $\pm$ 0.00e+00(-) & 15700.0000 & 15707.2333 $\pm$ 7.56e+00(=) & 15700.0000 & 15707.0000 $\pm$ 8.24e+00 \\ 
X-n148-k46 & \textbf{43448.0000} & 43448.0000 & 43448.0000 $\pm$ 0.00e+00(=) & 43448.0000 & 43448.0000 $\pm$ 0.00e+00(=) & 43448.0000 & 43448.0000 $\pm$ 0.00e+00 \\ 
X-n153-k22 & \textbf{21220.0000} & 21225.0000 & 21225.0000 $\pm$ 0.00e+00(+) & 21221.0000 & 21224.8667 $\pm$ 7.18e-01(+) & 21220.0000 & 21223.8000 $\pm$ 2.10e+00 \\ 
X-n157-k13 & \textbf{16876.0000} & 16876.0000 & 16876.0000 $\pm$ 0.00e+00(=) & 16876.0000 & 16876.0000 $\pm$ 0.00e+00(=) & 16876.0000 & 16876.0000 $\pm$ 0.00e+00 \\ 
X-n162-k11 & \textbf{14138.0000} & 14138.0000 & 14138.0000 $\pm$ 0.00e+00(=) & 14138.0000 & 14138.0000 $\pm$ 0.00e+00(=) & 14138.0000 & 14138.0000 $\pm$ 0.00e+00 \\ 
X-n167-k10 & \textbf{20557.0000} & 20557.0000 & 20557.0000 $\pm$ 0.00e+00(=) & 20557.0000 & 20557.0000 $\pm$ 0.00e+00(=) & 20557.0000 & 20557.0000 $\pm$ 0.00e+00 \\ 
X-n172-k51 & \textbf{45607.0000} & 45607.0000 & 45607.0000 $\pm$ 0.00e+00(=) & 45607.0000 & 45607.7000 $\pm$ 2.30e+00(=) & 45607.0000 & 45609.5333 $\pm$ 7.71e+00 \\ 
X-n176-k26 & \textbf{47812.0000} & 47812.0000 & 47812.0000 $\pm$ 0.00e+00(=) & 47812.0000 & 47812.0000 $\pm$ 0.00e+00(=) & 47812.0000 & 47812.0000 $\pm$ 0.00e+00 \\ 
X-n181-k23 & \textbf{25569.0000} & 25569.0000 & 25569.0000 $\pm$ 0.00e+00(-) & 25569.0000 & 25570.7000 $\pm$ 2.62e+00(=) & 25569.0000 & 25572.0000 $\pm$ 5.51e+00 \\ 
X-n186-k15 & \textbf{24145.0000} & 24145.0000 & 24145.0000 $\pm$ 0.00e+00(-) & 24145.0000 & 24149.7333 $\pm$ 3.69e+00(=) & 24145.0000 & 24150.6000 $\pm$ 3.48e+00 \\ 
X-n190-k8 & \textbf{16980.0000} & 16986.0000 & 16986.0000 $\pm$ 0.00e+00(+) & 16980.0000 & 16981.4000 $\pm$ 2.01e+00(=) & 16980.0000 & 16981.2333 $\pm$ 2.11e+00 \\ 
X-n195-k51 & \textbf{44225.0000} & 44225.0000 & 44225.0000 $\pm$ 0.00e+00(-) & 44225.0000 & 44257.3333 $\pm$ 1.79e+01(=) & 44234.0000 & 44259.7667 $\pm$ 1.13e+01 \\ 
X-n200-k36 & \textbf{58578.0000} & 58578.0000 & 58578.0000 $\pm$ 0.00e+00(-) & 58578.0000 & 58588.3333 $\pm$ 1.48e+01(=) & 58578.0000 & 58590.5000 $\pm$ 1.69e+01 \\ 
X-n204-k19 & \textbf{19565.0000} & 19565.0000 & 19565.0000 $\pm$ 0.00e+00(=) & 19565.0000 & 19566.4333 $\pm$ 4.21e+00(=) & 19565.0000 & 19566.3000 $\pm$ 4.12e+00 \\ 
X-n209-k16 & \textbf{30656.0000} & 30656.0000 & 30656.0000 $\pm$ 0.00e+00(-) & 30656.0000 & 30661.8667 $\pm$ 6.54e+00(=) & 30656.0000 & 30659.2000 $\pm$ 5.79e+00 \\ 
X-n214-k11 & \textbf{10856.0000} & 10856.0000 & 10856.0000 $\pm$ 0.00e+00(-) & 10867.0000 & 10869.4667 $\pm$ 2.38e+00(=) & 10867.0000 & 10869.6333 $\pm$ 4.36e+00 \\ 
X-n219-k73 & \textbf{117595.0000} & 117595.0000 & 117595.0000 $\pm$ 0.00e+00(=) & 117595.0000 & 117595.0000 $\pm$ 0.00e+00(=) & 117595.0000 & 117595.0000 $\pm$ 0.00e+00 \\ 
X-n223-k34 & \textbf{40437.0000} & 40437.0000 & 40437.0000 $\pm$ 0.00e+00(-) & 40437.0000 & 40442.7000 $\pm$ 1.36e+01(=) & 40437.0000 & 40451.8000 $\pm$ 2.31e+01 \\ 
X-n228-k23 & \textbf{25742.0000} & 25743.0000 & 25743.0000 $\pm$ 0.00e+00(=) & 25742.0000 & 25752.8000 $\pm$ 1.74e+01(+) & 25742.0000 & 25743.2667 $\pm$ 1.63e+00 \\ 
X-n233-k16 & \textbf{19230.0000} & 19230.0000 & 19230.0000 $\pm$ 0.00e+00(-) & 19230.0000 & 19256.1667 $\pm$ 1.95e+01(=) & 19230.0000 & 19246.8333 $\pm$ 2.53e+01 \\ 
X-n237-k14 & \textbf{27042.0000} & 27042.0000 & 27042.0000 $\pm$ 0.00e+00(-) & 27042.0000 & 27051.6333 $\pm$ 1.60e+01(=) & 27042.0000 & 27047.3333 $\pm$ 1.09e+01 \\ 
X-n242-k48 & \textbf{82751.0000} & 82789.0000 & 82790.5000 $\pm$ 1.50e+00(-) & 82751.0000 & 82796.6000 $\pm$ 4.90e+01(=) & 82751.0000 & 82813.1667 $\pm$ 4.21e+01 \\ 
X-n247-k50 & \textbf{37274.0000} & 37274.0000 & 37274.0000 $\pm$ 0.00e+00(=) & 37274.0000 & 37274.0667 $\pm$ 2.49e-01(=) & 37274.0000 & 37274.4000 $\pm$ 1.23e+00 \\ 
X-n251-k28 & \textbf{38684.0000} & 38699.0000 & 38699.0000 $\pm$ 0.00e+00(-) & 38684.0000 & 38740.6000 $\pm$ 3.47e+01(=) & 38684.0000 & 38744.9000 $\pm$ 3.88e+01 \\ 
X-n256-k16 & \textbf{18839.0000} & 18839.0000 & 18839.0000 $\pm$ 0.00e+00(-) & 18839.0000 & 18868.3667 $\pm$ 1.47e+01(=) & 18839.0000 & 18871.3000 $\pm$ 2.03e+01 \\ 
X-n261-k13 & \textbf{26558.0000} & 26558.0000 & 26558.0000 $\pm$ 0.00e+00(-) & 26558.0000 & 26572.1333 $\pm$ 1.67e+01(=) & 26558.0000 & 26581.8333 $\pm$ 2.02e+01 \\ 
X-n266-k58 & \textbf{75478.0000} & 75575.0000 & 75575.0000 $\pm$ 0.00e+00(+) & 75478.0000 & 75553.7333 $\pm$ 4.63e+01(=) & 75478.0000 & 75546.6333 $\pm$ 4.89e+01 \\ 
X-n270-k35 & \textbf{35291.0000} & 35303.0000 & 35303.0000 $\pm$ 0.00e+00(-) & 35303.0000 & 35323.3000 $\pm$ 2.03e+01(=) & 35303.0000 & 35319.4333 $\pm$ 1.90e+01 \\ 
X-n275-k28 & \textbf{21245.0000} & 21245.0000 & 21245.0000 $\pm$ 0.00e+00(=) & 21245.0000 & 21247.5333 $\pm$ 1.03e+01(=) & 21245.0000 & 21246.8333 $\pm$ 7.14e+00 \\ 
X-n280-k17 & \textbf{33503.0000} & 33506.0000 & 33506.0000 $\pm$ 0.00e+00(-) & 33505.0000 & 33565.3667 $\pm$ 2.93e+01(+) & 33505.0000 & 33549.5333 $\pm$ 2.82e+01 \\ 
X-n284-k15 & \textbf{20226.0000} & 20243.0000 & 20243.0000 $\pm$ 0.00e+00(-) & 20225.0000 & 20257.5000 $\pm$ 1.26e+01(=) & 20226.0000 & 20255.6000 $\pm$ 1.31e+01 \\ 
X-n289-k60 & \textbf{95151.0000} & 95323.0000 & 95323.0000 $\pm$ 0.00e+00(+) & 95180.0000 & 95252.4667 $\pm$ 4.24e+01(=) & 95151.0000 & 95248.9667 $\pm$ 4.45e+01 \\ 
X-n294-k50 & \textbf{47161.0000} & 47172.0000 & 47172.0000 $\pm$ 0.00e+00(-) & 47167.0000 & 47204.0000 $\pm$ 2.93e+01(=) & 47169.0000 & 47212.7333 $\pm$ 2.78e+01 \\ 
X-n298-k31 & \textbf{34231.0000} & 34231.0000 & 34231.0000 $\pm$ 0.00e+00(-) & 34231.0000 & 34261.0000 $\pm$ 1.73e+01(=) & 34231.0000 & 34262.5667 $\pm$ 2.30e+01 \\ 
X-n303-k21 & \textbf{21736.0000} & 21738.0000 & 21738.0000 $\pm$ 0.00e+00(-) & 21738.0000 & 21759.1667 $\pm$ 3.01e+01(=) & 21747.0000 & 21758.8333 $\pm$ 3.12e+01 \\ 
X-n308-k13 & \textbf{25859.0000} & 25862.0000 & 25862.0000 $\pm$ 0.00e+00(-) & 25861.0000 & 25880.8000 $\pm$ 2.19e+01(=) & 25861.0000 & 25871.9000 $\pm$ 2.01e+01 \\ 
X-n313-k71 & \textbf{94043.0000} & 94051.0000 & 94051.0000 $\pm$ 0.00e+00(-) & 94055.0000 & 94096.9333 $\pm$ 3.66e+01(=) & 94044.0000 & 94101.5000 $\pm$ 3.76e+01 \\ 
X-n317-k53 & \textbf{78355.0000} & 78368.0000 & 78368.0000 $\pm$ 0.00e+00(+) & 78355.0000 & 78355.5333 $\pm$ 1.36e+00(=) & 78355.0000 & 78355.6667 $\pm$ 1.49e+00 \\ 
X-n322-k28 & \textbf{29834.0000} & 29855.0000 & 29855.0000 $\pm$ 0.00e+00(-) & 29854.0000 & 29865.3000 $\pm$ 1.66e+01(=) & 29848.0000 & 29871.9333 $\pm$ 1.58e+01 \\ 
X-n327-k20 & \textbf{27532.0000} & 27557.0000 & 27557.0000 $\pm$ 0.00e+00(-) & 27547.0000 & 27583.8667 $\pm$ 2.01e+01(=) & 27532.0000 & 27588.3333 $\pm$ 2.23e+01 \\ 
X-n331-k15 & \textbf{31102.0000} & 31103.0000 & 31103.0000 $\pm$ 0.00e+00(-) & 31103.0000 & 31104.9333 $\pm$ 6.76e+00(=) & 31103.0000 & 31105.8000 $\pm$ 5.46e+00 \\ 
X-n336-k84 & \textbf{139111.0000} & 139272.0000 & 139272.0000 $\pm$ 0.00e+00(=) & 139139.0000 & 139283.5333 $\pm$ 7.05e+01(=) & 139197.0000 & 139269.6667 $\pm$ 4.81e+01 \\ 
X-n344-k43 & \textbf{42050.0000} & 42064.0000 & 42064.0000 $\pm$ 0.00e+00(-) & 42062.0000 & 42112.8333 $\pm$ 3.06e+01(+) & 42056.0000 & 42097.0000 $\pm$ 2.53e+01 \\ 
X-n351-k40 & \textbf{25896.0000} & 25955.0000 & 25955.0000 $\pm$ 0.00e+00(+) & 25927.0000 & 25954.1667 $\pm$ 2.07e+01(=) & 25925.0000 & 25948.0333 $\pm$ 1.41e+01 \\ 
X-n359-k29 & \textbf{51505.0000} & 51574.0000 & 51574.0000 $\pm$ 0.00e+00(+) & 51505.0000 & 51534.4667 $\pm$ 2.30e+01(-) & 51505.0000 & 51556.3333 $\pm$ 4.21e+01 \\ 
X-n367-k17 & \textbf{22814.0000} & 22814.0000 & 22814.0000 $\pm$ 0.00e+00(-) & 22814.0000 & 22822.3000 $\pm$ 5.60e+00(=) & 22814.0000 & 22824.6667 $\pm$ 5.22e+00 \\ 
X-n376-k94 & \textbf{147713.0000} & 147713.0000 & 147713.0000 $\pm$ 0.00e+00(-) & 147713.0000 & 147714.2667 $\pm$ 1.46e+00(=) & 147713.0000 & 147713.9333 $\pm$ 1.41e+00 \\ 
X-n384-k52 & \textbf{65940.0000} & 66074.0000 & 66074.0000 $\pm$ 0.00e+00(+) & 65966.0000 & 66024.5000 $\pm$ 4.43e+01(=) & 65939.0000 & 66001.3000 $\pm$ 4.58e+01 \\ 
X-n393-k38 & \textbf{38260.0000} & 38260.0000 & 38260.0000 $\pm$ 0.00e+00(-) & 38260.0000 & 38284.7667 $\pm$ 2.99e+01(=) & 38260.0000 & 38286.6333 $\pm$ 3.17e+01 \\ 
X-n401-k29 & \textbf{66154.0000} & 66213.0000 & 66213.0000 $\pm$ 0.00e+00(-) & 66186.0000 & 66216.7333 $\pm$ 1.63e+01(=) & 66193.0000 & 66219.5667 $\pm$ 1.15e+01 \\ 
X-n411-k19 & \textbf{19712.0000} & 19717.0000 & 19717.0000 $\pm$ 0.00e+00(-) & 19719.0000 & 19739.2667 $\pm$ 1.51e+01(+) & 19716.0000 & 19731.3667 $\pm$ 1.03e+01 \\ 
X-n420-k130 & \textbf{107798.0000} & 107888.0000 & 107888.0000 $\pm$ 0.00e+00(+) & 107798.0000 & 107823.3000 $\pm$ 1.70e+01(=) & 107798.0000 & 107822.0333 $\pm$ 2.19e+01 \\ 
X-n429-k61 & \textbf{65449.0000} & 65508.0000 & 65508.0000 $\pm$ 0.00e+00(-) & 65468.0000 & 65519.9333 $\pm$ 2.62e+01(=) & 65462.0000 & 65525.1000 $\pm$ 2.95e+01 \\ 
X-n439-k37 & \textbf{36391.0000} & 36395.0000 & 36395.0000 $\pm$ 0.00e+00(-) & 36395.0000 & 36407.7667 $\pm$ 1.18e+01(=) & 36394.0000 & 36402.8667 $\pm$ 1.06e+01 \\ 
X-n449-k29 & \textbf{55233.0000} & 55349.0000 & 55349.0000 $\pm$ 0.00e+00(+) & 55239.0000 & 55306.6333 $\pm$ 3.62e+01(=) & 55237.0000 & 55304.7000 $\pm$ 3.71e+01 \\ 
X-n459-k26 & \textbf{24139.0000} & 24160.0000 & 24160.0000 $\pm$ 0.00e+00(=) & 24141.0000 & 24162.0000 $\pm$ 1.56e+01(=) & 24139.0000 & 24162.8000 $\pm$ 1.51e+01 \\ 
X-n469-k138 & \textbf{221824.0000} & 222159.0000 & 222159.0000 $\pm$ 0.00e+00(+) & 221872.0000 & 222025.9333 $\pm$ 1.11e+02(=) & 221861.0000 & 222033.0667 $\pm$ 8.05e+01 \\ 
X-n480-k70 & \textbf{89449.0000} & 89468.0000 & 89468.0000 $\pm$ 0.00e+00(=) & 89449.0000 & 89461.6667 $\pm$ 9.69e+00(=) & 89449.0000 & 89469.0333 $\pm$ 1.78e+01 \\ 
X-n491-k59 & \textbf{66483.0000} & 66531.0000 & 66531.0000 $\pm$ 0.00e+00(-) & 66508.0000 & 66570.7000 $\pm$ 5.93e+01(=) & 66485.0000 & 66576.7000 $\pm$ 6.02e+01 \\ 
X-n502-k39 & \textbf{69226.0000} & 69268.0000 & 69268.0000 $\pm$ 0.00e+00(+) & 69226.0000 & 69231.3000 $\pm$ 8.87e+00(=) & 69226.0000 & 69234.9000 $\pm$ 8.32e+00 \\ 
X-n513-k21 & \textbf{24201.0000} & 24201.0000 & 24201.0000 $\pm$ 0.00e+00(-) & 24201.0000 & 24222.6000 $\pm$ 2.82e+01(+) & 24201.0000 & 24209.2667 $\pm$ 1.70e+01 \\ 
X-n524-k153 & \textbf{154593.0000} & 154755.0000 & 154755.0000 $\pm$ 0.00e+00(+) & 154602.0000 & 154628.5333 $\pm$ 4.04e+01(=) & 154599.0000 & 154641.5667 $\pm$ 5.87e+01 \\ 
X-n536-k96 & \textbf{94846.0000} & 95101.0000 & 95101.0000 $\pm$ 0.00e+00(+) & 94854.0000 & 94917.8000 $\pm$ 3.24e+01(=) & 94845.0000 & 94921.8333 $\pm$ 4.33e+01 \\ 
X-n548-k50 & \textbf{86700.0000} & 86792.0000 & 86792.0000 $\pm$ 0.00e+00(+) & 86704.0000 & 86739.2333 $\pm$ 2.24e+01(=) & 86704.0000 & 86728.5667 $\pm$ 1.96e+01 \\ 
X-n561-k42 & \textbf{42717.0000} & 42734.0000 & 42734.0000 $\pm$ 0.00e+00(-) & 42719.0000 & 42763.3333 $\pm$ 2.71e+01(+) & 42719.0000 & 42750.0000 $\pm$ 1.87e+01 \\ 
X-n573-k30 & \textbf{50673.0000} & 50798.0000 & 50814.7000 $\pm$ 1.17e+01(+) & 50720.0000 & 50736.8333 $\pm$ 1.40e+01(+) & 50720.0000 & 50730.5000 $\pm$ 9.07e+00 \\ 
X-n586-k159 & \textbf{190316.0000} & 190537.0000 & 190537.0000 $\pm$ 0.00e+00(+) & 190316.0000 & 190393.6333 $\pm$ 3.88e+01(=) & 190316.0000 & 190393.3667 $\pm$ 4.12e+01 \\ 
X-n599-k92 & \textbf{108451.0000} & 108617.0000 & 108635.8333 $\pm$ 1.91e+01(+) & 108475.0000 & 108577.7667 $\pm$ 4.99e+01(=) & 108484.0000 & 108568.5333 $\pm$ 5.41e+01 \\ 
X-n613-k62 & \textbf{59535.0000} & 59632.0000 & 59632.0000 $\pm$ 0.00e+00(+) & 59545.0000 & 59602.4000 $\pm$ 3.65e+01(=) & 59536.0000 & 59610.9333 $\pm$ 4.56e+01 \\ 
X-n627-k43 & \textbf{62164.0000} & 62384.0000 & 62389.1333 $\pm$ 2.69e+00(+) & 62170.0000 & 62222.2000 $\pm$ 3.35e+01(=) & 62175.0000 & 62210.0333 $\pm$ 2.64e+01 \\ 
X-n641-k35 & \textbf{63684.0000} & 63867.0000 & 63867.0000 $\pm$ 0.00e+00(+) & 63713.0000 & 63765.7667 $\pm$ 2.71e+01(=) & 63705.0000 & 63768.2333 $\pm$ 3.78e+01 \\ 
X-n655-k131 & \textbf{106780.0000} & 106818.0000 & 106818.0000 $\pm$ 0.00e+00(+) & 106780.0000 & 106788.8667 $\pm$ 1.01e+01(=) & 106780.0000 & 106790.1000 $\pm$ 9.82e+00 \\ 
X-n670-k130 & \textbf{146332.0000} & 146848.0000 & 146848.1000 $\pm$ 5.39e-01(=) & 146441.0000 & 146787.5667 $\pm$ 1.33e+02(=) & 146459.0000 & 146810.2000 $\pm$ 1.46e+02 \\ 
X-n685-k75 & \textbf{68205.0000} & 68344.0000 & 68344.0000 $\pm$ 0.00e+00(+) & 68227.0000 & 68299.7333 $\pm$ 4.37e+01(=) & 68227.0000 & 68290.9000 $\pm$ 3.92e+01 \\ 
X-n701-k44 & \textbf{81923.0000} & 82352.0000 & 82353.2333 $\pm$ 2.06e+00(+) & 81939.0000 & 81987.6000 $\pm$ 4.40e+01(=) & 81931.0000 & 81984.3333 $\pm$ 4.78e+01 \\ 
X-n716-k35 & \textbf{43373.0000} & 43489.0000 & 43489.0000 $\pm$ 0.00e+00(+) & 43371.0000 & 43402.8667 $\pm$ 2.30e+01(=) & 43371.0000 & 43392.8667 $\pm$ 2.16e+01 \\ 
\bottomrule
\end{tabular}
\begin{tablenotes}
\footnotesize
\item[a] Better BKS obtained by Ours.
\end{tablenotes}
\end{threeparttable}
}
\caption{Results achieves on the moderate-scale instances among compared methods.}
\label{tab:res_small}
\end{table*}

\begin{table*}[ht]
\centering
\resizebox{0.9\textwidth}{!}{
\begin{threeparttable}
\begin{tabular}{cccccccc}
\toprule
Instance & BKS & \multicolumn{2}{c}{HGS} & \multicolumn{2}{c}{AILS-II} & \multicolumn{2}{c}{Ours-EN} \\
\cmidrule(lr){3-4} \cmidrule(lr){5-6} \cmidrule(lr){7-8} \\
 &  & Best & Avg.(\%) (+/-/=) & Best & Avg.(\%) (+/-/=) & Best & Avg.(\%) \\
\midrule
X-n733-k159 & \textbf{136187.0000} & 136457.0000 & 136457.0000 $\pm$ 0.00e+00(+) & 136213.0000 & 136286.0667 $\pm$ 3.70e+01(=) & 136211.0000 & 136281.1333 $\pm$ 4.26e+01 \\ 
X-n749-k98 & \textbf{77269.0000} & 77796.0000 & 77798.5667 $\pm$ 4.65e+00(+) & 77334.0000 & 77395.9333 $\pm$ 3.78e+01(=) & 77335.0000 & 77405.4333 $\pm$ 4.56e+01 \\ 
X-n766-k71 & \textbf{114417.0000} & 114754.0000 & 114754.0000 $\pm$ 0.00e+00(+) & 114427.0000 & 114477.9333 $\pm$ 4.37e+01(=) & 114426.0000 & 114466.9000 $\pm$ 3.49e+01 \\ 
X-n783-k48 & \textbf{72386.0000} & 72759.0000 & 72759.0000 $\pm$ 0.00e+00(+) & 72423.0000 & 72482.0667 $\pm$ 4.73e+01(=) & 72427.0000 & 72480.4667 $\pm$ 3.83e+01 \\ 
X-n801-k40 & \textbf{73311.0000} & 73474.0000 & 73474.0000 $\pm$ 0.00e+00(+) & 73311.0000 & 73364.3000 $\pm$ 3.44e+01(=) & 73311.0000 & 73369.7000 $\pm$ 3.91e+01 \\ 
X-n819-k171 & \textbf{158121.0000} & 158491.0000 & 158647.8667 $\pm$ 6.74e+01(+) & 158199.0000 & 158257.1667 $\pm$ 3.26e+01(=) & 158176.0000 & 158247.0333 $\pm$ 3.18e+01 \\ 
X-n837-k142 & \textbf{193737.0000} & 194141.0000 & 194267.8667 $\pm$ 7.09e+01(+) & 193739.0000 & 193825.0667 $\pm$ 4.78e+01(=) & 193764.0000 & 193816.5000 $\pm$ 3.34e+01 \\ 
X-n856-k95 & \textbf{88965.0000} & 88990.0000 & 88990.0000 $\pm$ 0.00e+00(-) & 88966.0000 & 89012.8333 $\pm$ 3.21e+01(=) & 88966.0000 & 89014.1333 $\pm$ 2.75e+01 \\ 
X-n876-k59 & \textbf{99299.0000} & 99687.0000 & 99705.1667 $\pm$ 1.05e+01(+) & 99352.0000 & 99426.2333 $\pm$ 4.89e+01(=) & 99347.0000 & 99442.0333 $\pm$ 4.33e+01 \\ 
X-n895-k37 & \textbf{53860.0000} & 54098.0000 & 54098.0000 $\pm$ 0.00e+00(+) & 53855.0000 & 53952.5333 $\pm$ 5.62e+01(=) & 53860.0000 & 53931.3333 $\pm$ 4.13e+01 \\ 
X-n916-k207 & \textbf{329179.0000} & 329909.0000 & 329937.0333 $\pm$ 1.45e+01(+) & 329225.0000 & 329315.3000 $\pm$ 4.02e+01(=) & 329230.0000 & 329297.4667 $\pm$ 4.59e+01 \\ 
X-n936-k151 & \textbf{132715.0000} & 133599.0000 & 133612.6333 $\pm$ 2.73e+01(+) & 132919.0000 & 132971.4667 $\pm$ 3.87e+01(=) & 132880.0000 & 132983.8333 $\pm$ 4.63e+01 \\ 
X-n957-k87 & \textbf{85465.0000} & 85581.0000 & 85581.0000 $\pm$ 0.00e+00(+) & 85470.0000 & 85508.4000 $\pm$ 2.03e+01(+) & 85469.0000 & 85495.7000 $\pm$ 2.08e+01 \\ 
X-n979-k58 & \textbf{118976.0000} & 119317.0000 & 119394.5333 $\pm$ 5.84e+01(+) & 118973.0000 & 119004.7333 $\pm$ 3.03e+01(=) & 118975.0000 & 119005.6667 $\pm$ 2.12e+01 \\ 
X-n1001-k43 & \textbf{72355.0000} & 72590.0000 & 72594.8667 $\pm$ 7.44e+00(+) & 72364.0000 & 72442.7667 $\pm$ 3.90e+01(=) & 72368.0000 & 72426.9333 $\pm$ 3.94e+01 \\ 
\midrule 
Average & & 0.1056 & 0.109 (40/38/22) & 0.0152 & 0.0702 (9/1/90) & 0.0135 & 0.0672 \\ 

\bottomrule
\end{tabular}
\begin{tablenotes}
\footnotesize
\item[a] Better BKS obtained by Ours.
\end{tablenotes}
\end{threeparttable}
}
\caption{Results achieves on the moderate-scale instances among compared methods -- continue.}
\label{tab:res_100_2}
\end{table*}

\begin{table*}[ht]
\centering
\resizebox{0.9\textwidth}{!}{
\begin{threeparttable}
\begin{tabular}{cccccccc}
\toprule
Instance & BKS & \multicolumn{2}{c}{Ours-PFD} & \multicolumn{2}{c}{Ours-DDD} & \multicolumn{2}{c}{Ours-EN} \\
\cmidrule(lr){3-4} \cmidrule(lr){5-6} \cmidrule(lr){7-8} \\
 &  & Best & Avg.(\%) (+/-/=) & Best & Avg.(\%) (+/-/=) & Best & Avg.(\%) \\
\midrule 
X-n101-k25 & \textbf{27591.00} & 27591.00 & 27591.00 $\pm$ 0.00e+00(=) & 27591.00 & 27591.00 $\pm$ 0.00e+00(=) & 27591.00 & 27591.00 $\pm$ 0.00e+00 \\ 
X-n106-k14 & \textbf{26362.00} & 26362.00 & 26362.00 $\pm$ 0.00e+00(=) & 26362.00 & 26362.00 $\pm$ 0.00e+00(=) & 26362.00 & 26362.00 $\pm$ 0.00e+00 \\ 
X-n110-k13 & \textbf{14971.00} & 14971.00 & 14971.00 $\pm$ 0.00e+00(=) & 14971.00 & 14971.00 $\pm$ 0.00e+00(=) & 14971.00 & 14971.00 $\pm$ 0.00e+00 \\ 
X-n115-k10 & \textbf{12747.00} & 12747.00 & 12747.00 $\pm$ 0.00e+00(=) & 12747.00 & 12747.00 $\pm$ 0.00e+00(=) & 12747.00 & 12747.00 $\pm$ 0.00e+00 \\ 
X-n120-k6 & \textbf{13332.00} & 13332.00 & 13332.00 $\pm$ 0.00e+00(=) & 13332.00 & 13332.00 $\pm$ 0.00e+00(=) & 13332.00 & 13332.00 $\pm$ 0.00e+00 \\ 
X-n125-k30 & \textbf{55539.00} & 55539.00 & 55539.00 $\pm$ 0.00e+00(=) & 55539.00 & 55539.00 $\pm$ 0.00e+00(=) & 55539.00 & 55539.00 $\pm$ 0.00e+00 \\ 
X-n129-k18 & \textbf{28940.00} & 28940.00 & 28941.13 $\pm$ 3.17e+00(=) & 28940.00 & 28940.40 $\pm$ 1.58e+00(-) & 28940.00 & 28941.73 $\pm$ 3.21e+00 \\ 
X-n134-k13 & \textbf{10916.00} & 10916.00 & 10916.00 $\pm$ 0.00e+00(=) & 10916.00 & 10916.00 $\pm$ 0.00e+00(=) & 10916.00 & 10916.00 $\pm$ 0.00e+00 \\ 
X-n139-k10 & \textbf{13590.00} & 13590.00 & 13590.00 $\pm$ 0.00e+00(=) & 13590.00 & 13590.00 $\pm$ 0.00e+00(=) & 13590.00 & 13590.00 $\pm$ 0.00e+00 \\ 
X-n143-k7 & \textbf{15700.00} & 15700.00 & 15704.50 $\pm$ 7.60e+00(=) & 15700.00 & 15706.63 $\pm$ 7.94e+00(=) & 15700.00 & 15707.00 $\pm$ 8.24e+00 \\ 
X-n148-k46 & \textbf{43448.00} & 43448.00 & 43448.00 $\pm$ 0.00e+00(=) & 43448.00 & 43448.00 $\pm$ 0.00e+00(=) & 43448.00 & 43448.00 $\pm$ 0.00e+00 \\ 
X-n153-k22 & \textbf{21220.00} & 21220.00 & 21224.60 $\pm$ 1.96e+00(=) & 21220.00 & 21224.20 $\pm$ 1.80e+00(=) & 21220.00 & 21223.80 $\pm$ 2.10e+00 \\ 
X-n157-k13 & \textbf{16876.00} & 16876.00 & 16876.00 $\pm$ 0.00e+00(=) & 16876.00 & 16876.00 $\pm$ 0.00e+00(=) & 16876.00 & 16876.00 $\pm$ 0.00e+00 \\ 
X-n162-k11 & \textbf{14138.00} & 14138.00 & 14138.00 $\pm$ 0.00e+00(=) & 14138.00 & 14138.00 $\pm$ 0.00e+00(=) & 14138.00 & 14138.00 $\pm$ 0.00e+00 \\ 
X-n167-k10 & \textbf{20557.00} & 20557.00 & 20557.00 $\pm$ 0.00e+00(=) & 20557.00 & 20557.00 $\pm$ 0.00e+00(=) & 20557.00 & 20557.00 $\pm$ 0.00e+00 \\ 
X-n172-k51 & \textbf{45607.00} & 45607.00 & 45607.70 $\pm$ 3.77e+00(=) & 45607.00 & 45608.17 $\pm$ 5.43e+00(=) & 45607.00 & 45609.53 $\pm$ 7.71e+00 \\ 
X-n176-k26 & \textbf{47812.00} & 47812.00 & 47812.00 $\pm$ 0.00e+00(=) & 47812.00 & 47812.00 $\pm$ 0.00e+00(=) & 47812.00 & 47812.00 $\pm$ 0.00e+00 \\ 
X-n181-k23 & \textbf{25569.00} & 25569.00 & 25570.73 $\pm$ 1.61e+00(=) & 25569.00 & 25571.10 $\pm$ 2.90e+00(=) & 25569.00 & 25572.00 $\pm$ 5.51e+00 \\ 
X-n186-k15 & \textbf{24145.00} & 24145.00 & 24150.67 $\pm$ 3.52e+00(=) & 24145.00 & 24151.43 $\pm$ 3.53e+00(=) & 24145.00 & 24150.60 $\pm$ 3.48e+00 \\ 
X-n190-k8 & \textbf{16980.00} & 16980.00 & 16981.10 $\pm$ 2.77e+00(=) & 16980.00 & 16980.70 $\pm$ 1.46e+00(=) & 16980.00 & 16981.23 $\pm$ 2.11e+00 \\ 
X-n195-k51 & \textbf{44225.00} & 44225.00 & 44258.97 $\pm$ 1.55e+01(=) & 44225.00 & 44258.03 $\pm$ 1.38e+01(=) & 44234.00 & 44259.77 $\pm$ 1.13e+01 \\ 
X-n200-k36 & \textbf{58578.00} & 58578.00 & 58585.57 $\pm$ 1.40e+01(=) & 58578.00 & 58588.57 $\pm$ 1.55e+01(=) & 58578.00 & 58590.50 $\pm$ 1.69e+01 \\ 
X-n204-k19 & \textbf{19565.00} & 19565.00 & 19565.00 $\pm$ 0.00e+00(=) & 19565.00 & 19565.87 $\pm$ 3.33e+00(=) & 19565.00 & 19566.30 $\pm$ 4.12e+00 \\ 
X-n209-k16 & \textbf{30656.00} & 30656.00 & 30663.33 $\pm$ 7.39e+00(+) & 30656.00 & 30661.43 $\pm$ 6.33e+00(=) & 30656.00 & 30659.20 $\pm$ 5.79e+00 \\ 
X-n214-k11 & \textbf{10856.00} & 10863.00 & 10868.60 $\pm$ 2.20e+00(=) & 10867.00 & 10870.47 $\pm$ 3.81e+00(=) & 10867.00 & 10869.63 $\pm$ 4.36e+00 \\ 
X-n219-k73 & \textbf{117595.00} & 117595.00 & 117595.00 $\pm$ 0.00e+00(=) & 117595.00 & 117595.00 $\pm$ 0.00e+00(=) & 117595.00 & 117595.00 $\pm$ 0.00e+00 \\ 
X-n223-k34 & \textbf{40437.00} & 40437.00 & 40442.63 $\pm$ 3.53e+00(-) & 40437.00 & 40450.67 $\pm$ 2.10e+01(=) & 40437.00 & 40451.80 $\pm$ 2.31e+01 \\ 
X-n228-k23 & \textbf{25742.00} & 25742.00 & 25750.43 $\pm$ 1.69e+01(+) & 25742.00 & 25745.20 $\pm$ 8.37e+00(=) & 25742.00 & 25743.27 $\pm$ 1.63e+00 \\ 
X-n233-k16 & \textbf{19230.00} & 19230.00 & 19256.87 $\pm$ 2.57e+01(=) & 19230.00 & 19254.37 $\pm$ 2.57e+01(=) & 19230.00 & 19246.83 $\pm$ 2.53e+01 \\ 
X-n237-k14 & \textbf{27042.00} & 27042.00 & 27046.67 $\pm$ 8.27e+00(=) & 27042.00 & 27050.77 $\pm$ 1.43e+01(=) & 27042.00 & 27047.33 $\pm$ 1.09e+01 \\ 
X-n242-k48 & \textbf{82751.00} & 82751.00 & 82799.90 $\pm$ 4.35e+01(=) & 82751.00 & 82806.10 $\pm$ 4.96e+01(=) & 82751.00 & 82813.17 $\pm$ 4.21e+01 \\ 
X-n247-k50 & \textbf{37274.00} & 37274.00 & 37274.00 $\pm$ 0.00e+00(=) & 37274.00 & 37274.00 $\pm$ 0.00e+00(=) & 37274.00 & 37274.40 $\pm$ 1.23e+00 \\ 
X-n251-k28 & \textbf{38684.00} & 38684.00 & 38731.97 $\pm$ 3.65e+01(=) & 38684.00 & 38730.70 $\pm$ 3.66e+01(=) & 38684.00 & 38744.90 $\pm$ 3.88e+01 \\ 
X-n256-k16 & \textbf{18839.00} & 18839.00 & 18874.30 $\pm$ 1.12e+01(=) & 18839.00 & 18870.83 $\pm$ 1.57e+01(=) & 18839.00 & 18871.30 $\pm$ 2.03e+01 \\ 
X-n261-k13 & \textbf{26558.00} & 26558.00 & 26573.10 $\pm$ 1.71e+01(=) & 26558.00 & 26581.80 $\pm$ 2.04e+01(=) & 26558.00 & 26581.83 $\pm$ 2.02e+01 \\ 
X-n266-k58 & \textbf{75478.00} & 75478.00 & 75540.40 $\pm$ 4.59e+01(=) & 75478.00 & 75555.80 $\pm$ 3.78e+01(=) & 75478.00 & 75546.63 $\pm$ 4.89e+01 \\ 
X-n270-k35 & \textbf{35291.00} & 35303.00 & 35318.27 $\pm$ 1.95e+01(=) & 35303.00 & 35321.20 $\pm$ 1.66e+01(=) & 35303.00 & 35319.43 $\pm$ 1.90e+01 \\ 
X-n275-k28 & \textbf{21245.00} & 21245.00 & 21245.30 $\pm$ 1.62e+00(=) & 21245.00 & 21245.20 $\pm$ 1.08e+00(=) & 21245.00 & 21246.83 $\pm$ 7.14e+00 \\ 
X-n280-k17 & \textbf{33503.00} & 33505.00 & 33544.50 $\pm$ 2.73e+01(=) & 33505.00 & 33546.70 $\pm$ 2.51e+01(=) & 33505.00 & 33549.53 $\pm$ 2.82e+01 \\ 
X-n284-k15 & \textbf{20226.00} & 20225.00 & 20258.40 $\pm$ 1.17e+01(=) & 20241.00 & 20259.00 $\pm$ 1.07e+01(=) & 20226.00 & 20255.60 $\pm$ 1.31e+01 \\ 
X-n289-k60 & \textbf{95151.00} & 95151.00 & 95266.33 $\pm$ 3.93e+01(=) & 95175.00 & 95261.17 $\pm$ 3.95e+01(=) & 95151.00 & 95248.97 $\pm$ 4.45e+01 \\  
\bottomrule
\end{tabular}
\begin{tablenotes}
\footnotesize
\item[a] Better BKS obtained by Ours.
\end{tablenotes}
\end{threeparttable}
}
\caption{Comparison of results on moderate-scale instances among the top-3 designed heuristics.}
\label{tab:res_100_com_ours1}
\end{table*}

\begin{table*}[ht]
\centering
\resizebox{0.9\textwidth}{!}{
\begin{threeparttable}
\begin{tabular}{cccccccc}
\toprule
Instance & BKS & \multicolumn{2}{c}{Ours-PFD} & \multicolumn{2}{c}{Ours-DDD} & \multicolumn{2}{c}{Ours-EN} \\
\cmidrule(lr){3-4} \cmidrule(lr){5-6} \cmidrule(lr){7-8} \\
 &  & Best & Avg.(\%) (+/-/=) & Best & Avg.(\%) (+/-/=) & Best & Avg.(\%) \\
\midrule 
X-n294-k50 & \textbf{47161.00} & 47169.00 & 47209.33 $\pm$ 2.76e+01(=) & 47167.00 & 47201.50 $\pm$ 2.38e+01(=) & 47169.00 & 47212.73 $\pm$ 2.78e+01 \\ 
X-n298-k31 & \textbf{34231.00} & 34231.00 & 34262.40 $\pm$ 2.22e+01(=) & 34231.00 & 34250.97 $\pm$ 2.09e+01(-) & 34231.00 & 34262.57 $\pm$ 2.30e+01 \\
X-n303-k21 & \textbf{21736.00} & 21748.00 & 21767.17 $\pm$ 3.28e+01(=) & 21742.00 & 21763.77 $\pm$ 3.26e+01(=) & 21747.00 & 21758.83 $\pm$ 3.12e+01 \\ 
X-n308-k13 & \textbf{25859.00} & 25859.00 & 25871.93 $\pm$ 2.02e+01(=) & 25859.00 & 25873.40 $\pm$ 2.42e+01(=) & 25861.00 & 25871.90 $\pm$ 2.01e+01 \\ 
X-n313-k71 & \textbf{94043.00} & 94044.00 & 94089.33 $\pm$ 4.23e+01(=) & 94044.00 & 94089.57 $\pm$ 3.44e+01(=) & 94044.00 & 94101.50 $\pm$ 3.76e+01 \\ 
X-n317-k53 & \textbf{78355.00} & 78355.00 & 78356.03 $\pm$ 1.72e+00(=) & 78355.00 & 78355.67 $\pm$ 1.49e+00(=) & 78355.00 & 78355.67 $\pm$ 1.49e+00 \\ 
X-n322-k28 & \textbf{29834.00} & 29844.00 & 29864.17 $\pm$ 1.53e+01(=) & 29844.00 & 29871.20 $\pm$ 1.70e+01(=) & 29848.00 & 29871.93 $\pm$ 1.58e+01 \\ 
X-n327-k20 & \textbf{27532.00} & 27532.00 & 27586.77 $\pm$ 2.48e+01(=) & 27532.00 & 27586.17 $\pm$ 2.96e+01(=) & 27532.00 & 27588.33 $\pm$ 2.23e+01 \\ 
X-n331-k15 & \textbf{31102.00} & 31103.00 & 31104.17 $\pm$ 3.18e+00(=) & 31103.00 & 31103.90 $\pm$ 9.78e-01(=) & 31103.00 & 31105.80 $\pm$ 5.46e+00 \\
X-n336-k84 & \textbf{139111.00} & 139111.00 & 139264.50 $\pm$ 6.26e+01(=) & 139171.00 & 139287.90 $\pm$ 7.29e+01(=) & 139197.00 & 139269.67 $\pm$ 4.81e+01 \\ 
X-n344-k43 & \textbf{42050.00} & 42056.00 & 42104.20 $\pm$ 2.81e+01(=) & 42056.00 & 42094.77 $\pm$ 2.45e+01(=) & 42056.00 & 42097.00 $\pm$ 2.53e+01 \\ 
X-n351-k40 & \textbf{25896.00} & 25925.00 & 25951.37 $\pm$ 2.00e+01(=) & 25925.00 & 25955.30 $\pm$ 1.81e+01(=) & 25925.00 & 25948.03 $\pm$ 1.41e+01 \\ 
X-n359-k29 & \textbf{51505.00} & 51505.00 & 51558.57 $\pm$ 3.96e+01(=) & 51505.00 & 51539.17 $\pm$ 3.00e+01(=) & 51505.00 & 51556.33 $\pm$ 4.21e+01 \\ 
X-n367-k17 & \textbf{22814.00} & 22814.00 & 22824.33 $\pm$ 5.01e+00(=) & 22814.00 & 22825.33 $\pm$ 4.37e+00(=) & 22814.00 & 22824.67 $\pm$ 5.22e+00 \\ 
X-n376-k94 & \textbf{147713.00} & 147713.00 & 147714.00 $\pm$ 1.63e+00(=) & 147713.00 & 147713.80 $\pm$ 1.47e+00(=) & 147713.00 & 147713.93 $\pm$ 1.41e+00 \\ 
X-n384-k52 & \textbf{65940.00} & 65941.00 & 66026.80 $\pm$ 4.53e+01(+) & 65938.00 & 66009.77 $\pm$ 3.52e+01(=) & 65939.00 & 66001.30 $\pm$ 4.58e+01 \\ 
X-n393-k38 & \textbf{38260.00} & 38260.00 & 38278.53 $\pm$ 2.56e+01(=) & 38260.00 & 38278.80 $\pm$ 2.57e+01(=) & 38260.00 & 38286.63 $\pm$ 3.17e+01 \\ 
X-n401-k29 & \textbf{66154.00} & 66180.00 & 66211.23 $\pm$ 1.17e+01(-) & 66174.00 & 66214.43 $\pm$ 1.88e+01(=) & 66193.00 & 66219.57 $\pm$ 1.15e+01 \\ 
X-n411-k19 & \textbf{19712.00} & 19716.00 & 19729.60 $\pm$ 1.04e+01(=) & 19716.00 & 19733.90 $\pm$ 1.48e+01(=) & 19716.00 & 19731.37 $\pm$ 1.03e+01 \\ 
X-n420-k130 & \textbf{107798.00} & 107798.00 & 107822.70 $\pm$ 1.92e+01(=) & 107798.00 & 107829.87 $\pm$ 2.70e+01(=) & 107798.00 & 107822.03 $\pm$ 2.19e+01 \\ 
X-n429-k61 & \textbf{65449.00} & 65457.00 & 65527.70 $\pm$ 4.22e+01(=) & 65470.00 & 65532.50 $\pm$ 3.49e+01(=) & 65462.00 & 65525.10 $\pm$ 2.95e+01 \\ 
X-n439-k37 & \textbf{36391.00} & 36395.00 & 36403.23 $\pm$ 6.63e+00(=) & 36395.00 & 36404.03 $\pm$ 9.22e+00(=) & 36394.00 & 36402.87 $\pm$ 1.06e+01 \\ 
X-n449-k29 & \textbf{55233.00} & 55234.00 & 55293.47 $\pm$ 3.75e+01(=) & 55239.00 & 55308.43 $\pm$ 3.38e+01(=) & 55237.00 & 55304.70 $\pm$ 3.71e+01 \\ 
X-n459-k26 & \textbf{24139.00} & 24142.00 & 24170.43 $\pm$ 1.29e+01(+) & 24142.00 & 24171.60 $\pm$ 1.42e+01(+) & 24139.00 & 24162.80 $\pm$ 1.51e+01 \\ 
X-n469-k138 & \textbf{221824.00} & 221861.00 & 222014.73 $\pm$ 9.90e+01(=) & 221835.00 & 222023.70 $\pm$ 7.35e+01(=) & 221861.00 & 222033.07 $\pm$ 8.05e+01 \\ 
X-n480-k70 & \textbf{89449.00} & 89449.00 & 89460.83 $\pm$ 7.82e+00(-) & 89457.00 & 89467.83 $\pm$ 1.57e+01(=) & 89449.00 & 89469.03 $\pm$ 1.78e+01 \\ 
X-n491-k59 & \textbf{66483.00} & 66487.00 & 66564.93 $\pm$ 5.18e+01(=) & 66506.00 & 66568.50 $\pm$ 5.12e+01(=) & 66485.00 & 66576.70 $\pm$ 6.02e+01 \\ 
X-n502-k39 & \textbf{69226.00} & 69226.00 & 69236.90 $\pm$ 9.02e+00(=) & 69226.00 & 69235.43 $\pm$ 7.66e+00(=) & 69226.00 & 69234.90 $\pm$ 8.32e+00 \\ 
X-n513-k21 & \textbf{24201.00} & 24201.00 & 24227.67 $\pm$ 3.01e+01(+) & 24201.00 & 24218.73 $\pm$ 2.55e+01(=) & 24201.00 & 24209.27 $\pm$ 1.70e+01 \\ 
X-n524-k153 & \textbf{154593.00} & 154604.00 & 154635.57 $\pm$ 4.78e+01(=) & 154598.00 & 154627.20 $\pm$ 4.28e+01(=) & 154599.00 & 154641.57 $\pm$ 5.87e+01 \\ 
X-n536-k96 & \textbf{94846.00} & 94851.00 & 94924.67 $\pm$ 2.69e+01(=) & 94869.00 & 94923.70 $\pm$ 3.09e+01(=) & 94845.00 & 94921.83 $\pm$ 4.33e+01 \\ 
X-n548-k50 & \textbf{86700.00} & 86707.00 & 86731.13 $\pm$ 1.92e+01(=) & 86704.00 & 86734.13 $\pm$ 2.30e+01(=) & 86704.00 & 86728.57 $\pm$ 1.96e+01 \\ 
X-n561-k42 & \textbf{42717.00} & 42719.00 & 42762.10 $\pm$ 2.94e+01(=) & 42723.00 & 42752.13 $\pm$ 2.34e+01(=) & 42719.00 & 42750.00 $\pm$ 1.87e+01 \\ 
X-n573-k30 & \textbf{50673.00} & 50720.00 & 50736.63 $\pm$ 1.21e+01(+) & 50719.00 & 50730.10 $\pm$ 9.03e+00(=) & 50720.00 & 50730.50 $\pm$ 9.07e+00 \\ 
X-n586-k159 & \textbf{190316.00} & 190316.00 & 190401.73 $\pm$ 4.87e+01(=) & 190316.00 & 190396.70 $\pm$ 4.54e+01(=) & 190316.00 & 190393.37 $\pm$ 4.12e+01 \\ 
X-n599-k92 & \textbf{108451.00} & 108482.00 & 108571.67 $\pm$ 4.99e+01(=) & 108507.00 & 108589.93 $\pm$ 4.93e+01(=) & 108484.00 & 108568.53 $\pm$ 5.41e+01 \\ 
X-n613-k62 & \textbf{59535.00} & 59536.00 & 59592.53 $\pm$ 4.41e+01(=) & 59538.00 & 59604.23 $\pm$ 3.72e+01(=) & 59536.00 & 59610.93 $\pm$ 4.56e+01 \\ 
X-n627-k43 & \textbf{62164.00} & 62176.00 & 62199.40 $\pm$ 1.70e+01(=) & 62179.00 & 62205.80 $\pm$ 2.68e+01(=) & 62175.00 & 62210.03 $\pm$ 2.64e+01 \\ 
X-n641-k35 & \textbf{63684.00} & 63710.00 & 63760.53 $\pm$ 2.82e+01(=) & 63696.00 & 63766.53 $\pm$ 3.68e+01(=) & 63705.00 & 63768.23 $\pm$ 3.78e+01 \\ 
X-n655-k131 & \textbf{106780.00} & 106780.00 & 106789.40 $\pm$ 8.42e+00(=) & 106780.00 & 106793.53 $\pm$ 9.14e+00(=) & 106780.00 & 106790.10 $\pm$ 9.82e+00 \\ 
X-n670-k130 & \textbf{146332.00} & 146432.00 & 146781.10 $\pm$ 1.39e+02(=) & 146439.00 & 146806.97 $\pm$ 1.88e+02(=) & 146459.00 & 146810.20 $\pm$ 1.46e+02 \\ 
X-n685-k75 & \textbf{68205.00} & 68226.00 & 68272.83 $\pm$ 3.15e+01(=) & 68231.00 & 68293.23 $\pm$ 4.44e+01(=) & 68227.00 & 68290.90 $\pm$ 3.92e+01 \\ 
X-n701-k44 & \textbf{81923.00} & 81941.00 & 81983.03 $\pm$ 3.65e+01(=) & 81942.00 & 81985.30 $\pm$ 3.38e+01(=) & 81931.00 & 81984.33 $\pm$ 4.78e+01 \\ 
X-n716-k35 & \textbf{43373.00} & 43372.00 & 43398.83 $\pm$ 2.23e+01(=) & 43372.00 & 43384.33 $\pm$ 1.07e+01(=) & 43371.00 & 43392.87 $\pm$ 2.16e+01 \\
X-n733-k159 & \textbf{136187.00} & 136216.00 & 136277.20 $\pm$ 3.43e+01(=) & 136217.00 & 136274.33 $\pm$ 3.59e+01(=) & 136211.00 & 136281.13 $\pm$ 4.26e+01 \\ 
X-n749-k98 & \textbf{77269.00} & 77323.00 & 77421.07 $\pm$ 4.64e+01(=) & 77334.00 & 77415.73 $\pm$ 5.31e+01(=) & 77335.00 & 77405.43 $\pm$ 4.56e+01 \\ 
X-n766-k71 & \textbf{114417.00} & 114419.00 & 114473.03 $\pm$ 6.60e+01(=) & 114425.00 & 114466.30 $\pm$ 3.58e+01(=) & 114426.00 & 114466.90 $\pm$ 3.49e+01 \\ 
X-n783-k48 & \textbf{72386.00} & 72411.00 & 72479.50 $\pm$ 4.65e+01(=) & 72411.00 & 72494.30 $\pm$ 4.34e+01(=) & 72427.00 & 72480.47 $\pm$ 3.83e+01 \\ 
X-n801-k40 & 73311.00 & \textbf{73310.00} & 73350.60 $\pm$ 3.34e+01(=) & 73310.00 & 73361.07 $\pm$ 3.46e+01(=) & 73311.00 & 73369.70 $\pm$ 3.91e+01 \\ 
X-n819-k171 & \textbf{158121.00} & 158215.00 & 158262.53 $\pm$ 3.79e+01(=) & 158180.00 & 158245.73 $\pm$ 3.79e+01(=) & 158176.00 & 158247.03 $\pm$ 3.18e+01 \\ 
X-n837-k142 & \textbf{193737.00} & 193737.00 & 193800.50 $\pm$ 3.56e+01(=) & 193760.00 & 193820.13 $\pm$ 4.37e+01(=) & 193764.00 & 193816.50 $\pm$ 3.34e+01 \\ 
X-n856-k95 & \textbf{88965.00} & 88966.00 & 89017.53 $\pm$ 3.00e+01(=) & 88966.00 & 89015.40 $\pm$ 2.60e+01(=) & 88966.00 & 89014.13 $\pm$ 2.75e+01 \\ 
X-n876-k59 & \textbf{99299.00} & 99359.00 & 99432.70 $\pm$ 3.58e+01(=) & 99347.00 & 99413.93 $\pm$ 3.75e+01(-) & 99347.00 & 99442.03 $\pm$ 4.33e+01 \\ 
X-n895-k37 & \textbf{53860.00} & 53871.00 & 53968.13 $\pm$ 6.39e+01(+) & 53861.00 & 53960.70 $\pm$ 5.41e+01(+) & 53860.00 & 53931.33 $\pm$ 4.13e+01 \\ 
X-n916-k207 & \textbf{329179.00} & 329219.00 & 329297.43 $\pm$ 5.20e+01(=) & 329217.00 & 329289.40 $\pm$ 5.11e+01(=) & 329230.00 & 329297.47 $\pm$ 4.59e+01 \\ 
X-n936-k151 & \textbf{132715.00} & 132809.00 & 133000.43 $\pm$ 5.70e+01(=) & 132900.00 & 132991.07 $\pm$ 4.38e+01(=) & 132880.00 & 132983.83 $\pm$ 4.63e+01 \\ 
X-n957-k87 & \textbf{85465.00} & 85465.00 & 85497.90 $\pm$ 1.87e+01(=) & 85467.00 & 85505.03 $\pm$ 2.39e+01(=) & 85469.00 & 85495.70 $\pm$ 2.08e+01 \\ 
X-n979-k58 & 118976.00 & \textbf{118973.00} & 118999.50 $\pm$ 2.00e+01(=) & 118974.00 & 119011.83 $\pm$ 3.16e+01(=) & 118975.00 & 119005.67 $\pm$ 2.12e+01 \\ 
X-n1001-k43 & 72355.00 & \textbf{72353.00} & 72434.37 $\pm$ 4.49e+01(=) & 72376.00 & 72437.80 $\pm$ 4.03e+01(=) & 72368.00 & 72426.93 $\pm$ 3.94e+01 \\ 
\midrule 
Average & & 0.0113 & 0.0678 (7/3/90) & 0.0145 & 0.0686 (2/3/95) & 0.0135 & 0.0672  \\ 

\bottomrule
\end{tabular}
\end{threeparttable}
}
\caption{Comparison of Results on Moderate-Scale Instances Among the Top-3 Designed Heuristics -- continue.}
\label{tab:res_100_com_ours2}
\end{table*}

\clearpage

\section{New BKS Results}\label{app:new_bks}

In this section, we present the new best-known solutions (BKS) achieved by our designed heuristics. For large-scale instances, the new BKS is obtained in $10 * n$ seconds, where $n$ is the number of nodes. For moderate-scale instances, the BKS is achieved in $3 * n$ seconds. First, we present the convergence curves and corresponding solution quality in \sectionautorefname~\ref{subsec:ccnnbs}. Then, we detail the routes of these solutions in \sectionautorefname~\ref{subsec:nbr}.

\subsection{Convergence Curves and New Best-Known Solutions}\label{subsec:ccnnbs}

\begin{figure*}[ht]
    \centering
    \includegraphics[width=\textwidth]{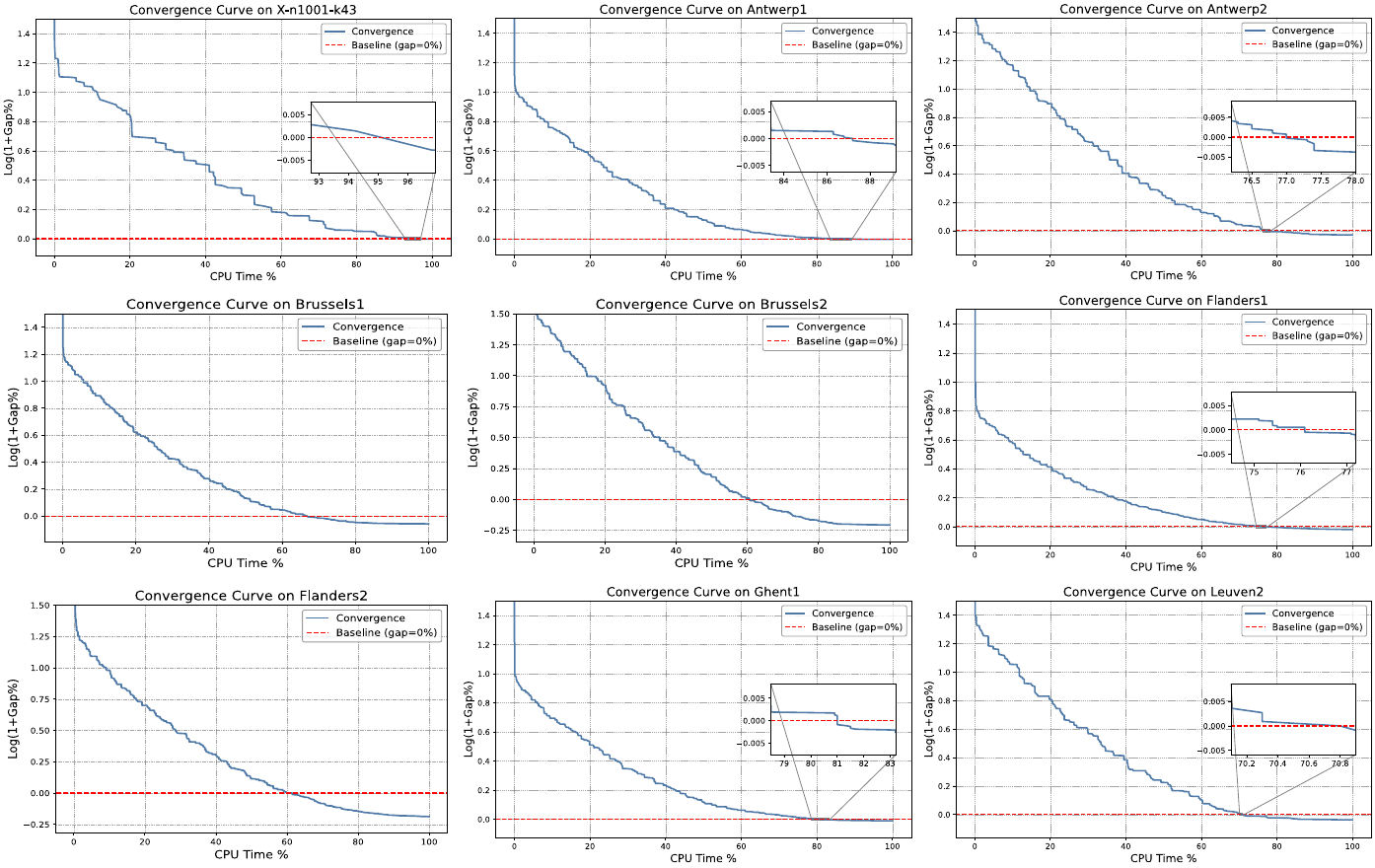}
    \caption{Convergence curves of the new best-known solutions. The x-axis represents the CPU time percentage, while the y-axis shows $\log(1 + \text{Gap\%})$. The red dashed line indicates the baseline (gap = 0\%).}
    \label{fig:bks_con}
\end{figure*}

Figure~\ref{fig:bks_con} illustrates the convergence curves of the new BKS achieved by our designed heuristics. Specifically, we obtain \textbf{1 new BKS} for the moderate-scale instance and \textbf{8 out of 10 new BKS} for large-scale instances.

For the moderate-scale instance X-n1001-k43, our method demonstrated consistent improvements throughout the optimization process, ultimately achieving a final result of 72353, matching the baseline but with slight refinements.

In large-scale instances, our method consistently outperformed the baseline, achieving new BKS for Antwerp1, Antwerp2, Brussels1, Brussels2, Flanders1, Flanders2, Ghent1, and Leuven2. The convergence curves reveal a trend of rapid improvement during the early stages of optimization, followed by steady refinements in later stages. Notably, the final gaps achieved by our method were consistently close to 0.25\% for Brussels2 and Flanders2, representing significant improvements over the baseline.

For Antwerp1, Antwerp2, Flanders1, Ghent1, and Leuven2, the gap reduced sharply within the first 50\% of CPU time, eventually converging to slightly better results than the baseline. A zoom-in box highlights the changes exceeding the baseline for clarity. For Brussels1, the final improvement was approximately 0.1\%.

\begin{figure*}[ht]
    \centering
    \includegraphics[width=\textwidth]{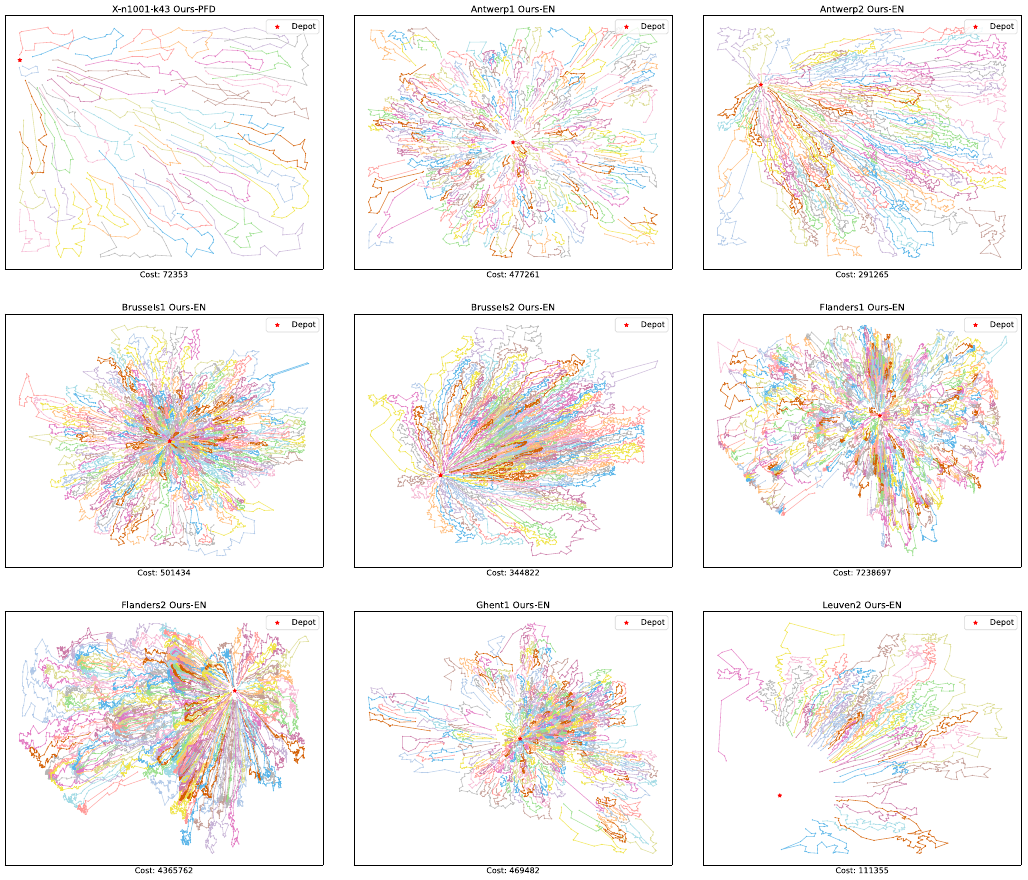}
    \caption{Visualization of the vehicle routes for the new best-known solutions.}
    \label{fig:bks_sol}
\end{figure*}

We also provide a detailed visualization of the specific routes for each new best-known solution, as shown in Figure~\ref{fig:bks_sol}.

\subsection{New BKS Routes}\label{subsec:nbr}

In this subsection, we provide the detailed routes for each new best-known solution. For a comprehensive overview, the complete route details can be found in the supplementary material.

\section{License for Used Resources}
\begin{table}[htbp]
\centering
\caption{List of licenses for the codes and datasets we used in this work}
\vspace{8pt}
\label{table:Licenses}
\resizebox{0.99\textwidth}{!}{%
\begin{tabular}{l |l | l | l }
\toprule
 Resource   &   Type  &  Link  & License    \\
\midrule
HGS \citep{vidal2022hybrid}& Code & \url{https://github.com/chkwon/PyHygese} & MIT License\\
AILS-II \citep{maximo2024ails} & Code & \url{https://github.com/vinymax10/AILS-CVRP} & MIT License \\
\midrule
CVRPLIB Moderate-scale \citep{uchoa2017new} & Dataset & \url{http://vrp.galgos.inf.puc-rio.br/index.php/en/} & Available for academic research use \\

CVRPLIB Large-scale \citep{arnold2019efficiently} & Dataset & \url{http://vrp.galgos.inf.puc-rio.br/index.php/en/}  & Available for academic research use \\
\bottomrule
\end{tabular}%
}
\end{table}
We list the used existing codes and datasets in \ref{table:Licenses}, and all of them are open-sourced resources for academic usage.

\section{Broader Impacts}
\label{append:broader_impacts}
This paper presents work whose goal is to advance the CVRP solver based on LLM. We believe the proposed AILS-AHD framework is valuable for promoting the further development of the field, such as improved efficiency in solving large-scale CVRP instances. This work can inspire follow-up works to explore more efficient LLM-driven methods for enhancing the solvers for routing problems. The proposed method has no potential negative societal impacts that we feel must be specifically highlighted.


\clearpage
{\small

\bibliographystyle{unsrtnat} 
\bibliography{myreference}        

@article{ails2022,
  title={An adaptive iterated local search heuristic for the heterogeneous fleet vehicle routing problem},
  author={M{\'a}ximo, Vin{\'\i}cius R and Cordeau, Jean-Fran{\c{c}}ois and Nascimento, Mari{\'a} CV},
  journal={Computers \& Operations Research},
  volume={148},
  pages={105954},
  year={2022},
  publisher={Elsevier}
}

@article{minaee2024large,
  title={Large language models: A survey},
  author={Minaee, Shervin and Mikolov, Tomas and Nikzad, Narjes and Chenaghlu, Meysam and Socher, Richard and Amatriain, Xavier and Gao, Jianfeng},
  journal={arXiv preprint arXiv:2402.06196},
  year={2024}
}

@article{jiang2024unco,
  title={Unco: Towards unifying neural combinatorial optimization through large language model},
  author={Jiang, Xia and Wu, Yaoxin and Wang, Yuan and Zhang, Yingqian},
  journal={arXiv preprint arXiv:2408.12214},
  year={2024}
}

@article{uhgs2014a,
  title={A unified solution framework for multi-attribute vehicle routing problems},
  author={Vidal, Thibaut and Crainic, Teodor Gabriel and Gendreau, Michel and Prins, Christian},
  journal={European Journal of Operational Research},
  volume={234},
  number={3},
  pages={658--673},
  year={2014},
  publisher={Elsevier}
}

@article{baldacci2010exact,
  title={An exact solution framework for a broad class of vehicle routing problems},
  author={Baldacci, Roberto and Bartolini, Enrico and Mingozzi, Aristide and Roberti, Roberto},
  journal={Computational Management Science},
  volume={7},
  pages={229--268},
  year={2010},
  publisher={Springer}
}

@article{laporte1992vehicle,
  title={The vehicle routing problem: An overview of exact and approximate algorithms},
  author={Laporte, Gilbert},
  journal={European journal of operational research},
  volume={59},
  number={3},
  pages={345--358},
  year={1992},
  publisher={Elsevier}
}

@article{schrimpf2000rr,
  title={Record breaking optimization results using the ruin and recreate principle},
  author={Schrimpf, Gerhard and Schneider, Johannes and Stamm-Wilbrandt, Hermann and Dueck, Gunter},
  journal={Journal of Computational Physics},
  volume={159},
  number={2},
  pages={139--171},
  year={2000},
  publisher={Elsevier}
}

@article{romera2024funsearch,
  title={Mathematical discoveries from program search with large language models},
  author={Romera-Paredes, Bernardino and Barekatain, Mohammadamin and Novikov, Alexander and Balog, Matej and Kumar, M Pawan and Dupont, Emilien and Ruiz, Francisco JR and Ellenberg, Jordan S and Wang, Pengming and Fawzi, Omar and others},
  journal={Nature},
  volume={625},
  number={7995},
  pages={468--475},
  year={2024},
  publisher={Nature Publishing Group UK London}
}

@inproceedings{Liu2024EvolutionOH,
  title={Evolution of Heuristics: Towards Efficient Automatic Algorithm Design Using Large Language Model},
  author={Liu, Fei and Xialiang, Tong and Yuan, Mingxuan and Lin, Xi and Luo, Fu and Wang, Zhenkun and Lu, Zhichao and Zhang, Qingfu},
  booktitle={Forty-first International Conference on Machine Learning},
  year={2024}
}

@article{ye2024reevo,
  title={ReEvo: Large Language Models as Hyper-Heuristics with Reflective Evolution},
  author={Ye, Haoran and Wang, Jiarui and Cao, Zhiguang and Song, Guojie},
  journal={arXiv preprint arXiv:2402.01145},
  year={2024}
}

@article{eiben2015evolutionary,
  title={From evolutionary computation to the evolution of things},
  author={Eiben, Agoston E and Smith, Jim},
  journal={Nature},
  volume={521},
  number={7553},
  pages={476--482},
  year={2015},
  publisher={Nature Publishing Group}
}

@article{boyd2007branch,
  title={Branch and bound methods},
  author={Boyd, Stephen and Mattingley, Jacob},
  journal={Notes for EE364b, Stanford University},
  volume={2006},
  pages={07},
  year={2007},
  publisher={Citeseer}
}

@article{subramanian2013hybrid,
  title={A hybrid algorithm for a class of vehicle routing problems},
  author={Subramanian, Anand and Uchoa, Eduardo and Ochi, Luiz Satoru},
  journal={Computers \& Operations Research},
  volume={40},
  number={10},
  pages={2519--2531},
  year={2013},
  publisher={Elsevier}
}

@article{maximo2021hybrid,
  title={A hybrid adaptive iterated local search with diversification control to the capacitated vehicle routing problem},
  author={M{\'a}ximo, Vin{\'\i}cius R and Nascimento, Mari{\'a} CV},
  journal={European Journal of Operational Research},
  volume={294},
  number={3},
  pages={1108--1119},
  year={2021},
  publisher={Elsevier}
}

@article{vidal2022hybrid,
  title={Hybrid genetic search for the CVRP: Open-source implementation and SWAP* neighborhood},
  author={Vidal, Thibaut},
  journal={Computers \& Operations Research},
  volume={140},
  pages={105643},
  year={2022},
  publisher={Elsevier}
}

@article{christiaens2020slack,
  title={Slack induction by string removals for vehicle routing problems},
  author={Christiaens, Jan and Vanden Berghe, Greet},
  journal={Transportation Science},
  volume={54},
  number={2},
  pages={417--433},
  year={2020},
  publisher={INFORMS}
}

@article{maximo2024ails,
  title={AILS-II: An adaptive iterated local search heuristic for the large-scale capacitated vehicle routing problem},
  author={M{\'a}ximo, Vin{\'\i}cius R and Cordeau, Jean-Fran{\c{c}}ois and Nascimento, Mari{\'a} CV},
  journal={INFORMS Journal on Computing},
  volume={36},
  number={4},
  pages={974--986},
  year={2024},
  publisher={INFORMS}
}

@incollection{lehman2023evolution,
  title={Evolution through large models},
  author={Lehman, Joel and Gordon, Jonathan and Jain, Shawn and Ndousse, Kamal and Yeh, Cathy and Stanley, Kenneth O},
  booktitle={Handbook of Evolutionary Machine Learning},
  pages={331--366},
  year={2023},
  publisher={Springer}
}

@article{meyerson2024language,
  title={Language model crossover: Variation through few-shot prompting},
  author={Meyerson, Elliot and Nelson, Mark J and Bradley, Herbie and Gaier, Adam and Moradi, Arash and Hoover, Amy K and Lehman, Joel},
  journal={ACM Transactions on Evolutionary Learning},
  volume={4},
  number={4},
  pages={1--40},
  year={2024},
  publisher={ACM New York, NY}
}

@article{accorsi2021fast,
  title={A fast and scalable heuristic for the solution of large-scale capacitated vehicle routing problems},
  author={Accorsi, Luca and Vigo, Daniele},
  journal={Transportation Science},
  volume={55},
  number={4},
  pages={832--856},
  year={2021},
  publisher={INFORMS}
}

@article{uchoa2017new,
  title={New benchmark instances for the capacitated vehicle routing problem},
  author={Uchoa, Eduardo and Pecin, Diego and Pessoa, Artur and Poggi, Marcus and Vidal, Thibaut and Subramanian, Anand},
  journal={European Journal of Operational Research},
  volume={257},
  number={3},
  pages={845--858},
  year={2017},
  publisher={Elsevier}
}

@article{konstantakopoulos2022vehicleapplication,
  title={Vehicle routing problem and related algorithms for logistics distribution: A literature review and classification},
  author={Konstantakopoulos, Grigorios D and Gayialis, Sotiris P and Kechagias, Evripidis P},
  journal={Operational research},
  volume={22},
  number={3},
  pages={2033--2062},
  year={2022},
  publisher={Springer}
}

@article{arnold2019efficiently,
  title={Efficiently solving very large-scale routing problems},
  author={Arnold, Florian and Gendreau, Michel and S{\"o}rensen, Kenneth},
  journal={Computers \& operations research},
  volume={107},
  pages={32--42},
  year={2019},
  publisher={Elsevier}
}

@article{potvin1995exchange2opts,
  title={An exchange heuristic for routeing problems with time windows},
  author={Potvin, Jean-Yves and Rousseau, Jean-Marc},
  journal={Journal of the Operational Research Society},
  volume={46},
  number={12},
  pages={1433--1446},
  year={1995},
  publisher={Taylor \& Francis}
}

@article{osman1993metastrategyswap,
  title={Metastrategy simulated annealing and tabu search algorithms for the vehicle routing problem},
  author={Osman, Ibrahim Hassan},
  journal={Annals of operations research},
  volume={41},
  pages={421--451},
  year={1993},
  publisher={Springer}
}

@article{lin1965computer2opt,
  title={Computer solutions of the traveling salesman problem},
  author={Lin, Shen},
  journal={Bell System Technical Journal},
  volume={44},
  number={10},
  pages={2245--2269},
  year={1965},
  publisher={Wiley Online Library}
}

@article{osman1993metastrategyshift,
  title={Metastrategy simulated annealing and tabu search algorithms for the vehicle routing problem},
  author={Osman, Ibrahim Hassan},
  journal={Annals of operations research},
  volume={41},
  pages={421--451},
  year={1993},
  publisher={Springer}
}

@article{wei2022chain,
  title={Chain-of-thought prompting elicits reasoning in large language models},
  author={Wei, Jason and Wang, Xuezhi and Schuurmans, Dale and Bosma, Maarten and Xia, Fei and Chi, Ed and Le, Quoc V and Zhou, Denny and others},
  journal={Advances in neural information processing systems},
  volume={35},
  pages={24824--24837},
  year={2022}
}
}

\clearpage


\clearpage

\end{document}